\documentclass{article}

\usepackage[utf8]{inputenc} 
\usepackage{url}            
\usepackage{booktabs}       
\usepackage{amsfonts}       
\usepackage{nicefrac}       
\usepackage{microtype}      
\usepackage{pifont}
\usepackage{comment}
\usepackage{epsfig}
\usepackage{graphics}
\usepackage{graphicx}
\usepackage{amsmath}
\usepackage{amssymb}
\usepackage{xspace}
\usepackage{bbm}
\usepackage{multirow}
\usepackage{enumitem}
\usepackage{epsf}
\usepackage{changepage}
\usepackage{hyperref}       
\usepackage[normalem]{ulem}
\usepackage[ruled,vlined,linesnumbered]{algorithm2e}

\usepackage[margin=1.4in]{geometry}
\usepackage{times}
\usepackage{authblk}
\usepackage{xcolor}
\hypersetup{
    colorlinks = true,
    linkcolor = red,
    anchorcolor = blue,
    citecolor = green,
    filecolor = blue,
    urlcolor = purple
}

\makeatletter
\renewcommand\paragraph{\@startsection{paragraph}{4}{\z@}%
    {0.5ex \@plus1ex \@minus.2ex}%
    {-0.5em}%
    {\normalfont\normalsize\bfseries}}
\makeatother

\def \ours{LOST\xspace}  
\def \cad{CAD\xspace}  
\def \od{OD\xspace}  
\def \cls{\texttt{CLS}\xspace}  

\title{Localizing Objects with Self-Supervised Transformers\\and no Labels} 
\date{}

\author[1]{Oriane Siméoni}
\author[1]{Gilles Puy}
\author[1,2]{Huy V. Vo}
\author[1,3]{Simon Roburin}
\author[1]{Spyros Gidaris}
\author[1]{Andrei Bursuc}
\author[1]{Patrick Pérez}
\author[1,3]{Renaud Marlet}
\author[2,4]{Jean Ponce}

\affil[1]{Valeo.ai, Paris, France}
\affil[2]{Inria and DIENS (ENS-PSL, CNRS, Inria), Paris, France}
\affil[3]{LIGM, Ecole des Ponts, Univ Gustave Eiffel, CNRS, Marne-la-Vall\'ee, France}
\affil[4]{Center for Data Science, New York University, New York, USA}

\newcommand{\red}[1]{{\color{red}{#1}}}

\newcommand{\citeme}[1]{\red{[XX]}}
\newcommand{\refme}[1]{\red{(XX)}}

\newcommand{\tran}{^\top}

\newcommand{\real}{\mathbb{R}}

\DeclareMathOperator*{\argmin}{arg\,min}

\newcommand{\cS}{\mathcal{S}}

\newcommand{\vA}{\mathbf{A}}

\newcommand{\vF}{\mathbf{F}}

\newcommand{\vI}{\mathbf{I}}

\newcommand{\vK}{\mathbf{K}}

\newcommand{\vQ}{\mathbf{Q}}

\newcommand{\vV}{\mathbf{V}}

\newcommand{\vX}{\mathbf{X}}
\newcommand{\vY}{\mathbf{Y}}

\newcommand{\vf}{\mathbf{f}}

\newcommand{\vk}{\mathbf{k}}

\newcommand{\vm}{\mathbf{m}}

\newcommand{\vq}{\mathbf{q}}

\newcommand{\vv}{\mathbf{v}}

\makeatletter
\newcommand*\bdot{\mathpalette\bdot@{.7}}
\newcommand*\bdot@[2]{\mathbin{\vcenter{\hbox{\scalebox{#2}{$\m@th#1\bullet$}}}}}
\makeatother

\makeatletter
\DeclareRobustCommand\onedot{\futurelet\@let@token\@onedot}
\def\@onedot{\ifx\@let@token.\else.\null\fi\xspace}
\def\eg{\emph{e.g}\onedot} 
\def\ie{\emph{i.e}\onedot}

\def\wrt{w.r.t\onedot}  
\def\etal{\emph{et al}\onedot}
\makeatother

\begin{document}

\maketitle

\begin{abstract}
   \noindent Localizing objects in image collections  without supervision can help to avoid expensive annotation campaigns. We propose a simple approach to this problem, that leverages the activation features of a vision transformer pre-trained in a self-supervised manner. Our method, \ours, does not require any external object proposal nor any exploration of the image collection; it operates on a single image. Yet, we outperform state-of-the-art object discovery methods by up to 8 CorLoc points on PASCAL VOC 2012. We also show that training a class-agnostic detector on the discovered objects boosts results by another 7 points. Moreover, we show promising results on the unsupervised object discovery task. The code to reproduce our results can be found at \url{https://github.com/valeoai/LOST}.
\end{abstract}

\begin{figure*}[h]
\centering\small
\includegraphics[width=\linewidth]{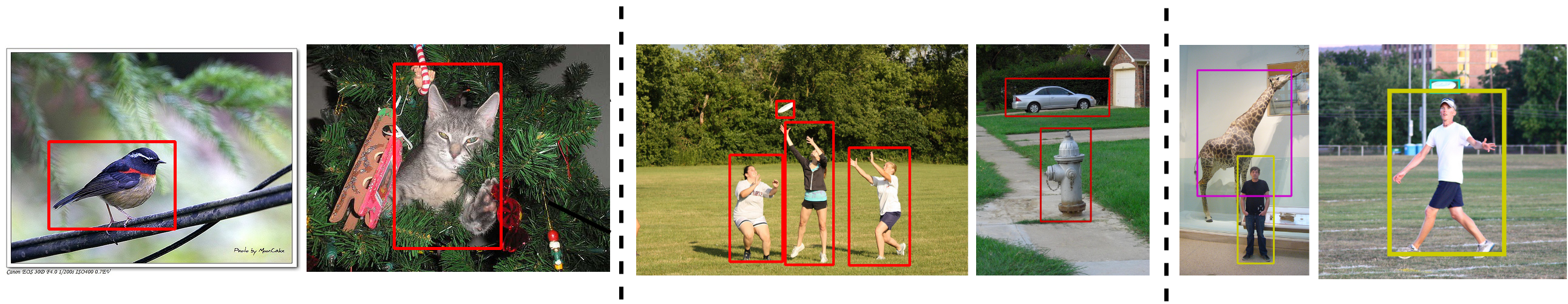}
\vspace*{-22pt}
\caption{\small 
    Three applications of LOST to unsupervised single-object discovery (left), multi-object discovery (middle) and object detection (right). In the latter case, objects discovered by \ours are clustered into categories, and cluster labels are used to train a classical object detector. Although large image collections are used to train the underlying image representation \cite{caron2021emerging} and the detector \cite{shaoqing2015faster}, {\em no annotation} is ever used in the pipeline. See \autoref{fig:similars} and {\color{red} Tables} \ref{tab:training}, \ref{tab:training-class-ap} for more experiments\rlap. 
}
\label{fig:teaser}
\end{figure*} 

\vspace{-20pt}

\section{Introduction}

Object detectors are now part of critical systems, such as autonomous vehicles. However, to reach a high level of performance, they are trained on a vast amount of costly annotated data. Various approaches have been proposed to reduce these costs, such as semi-supervision \cite{liu2021unbiased}, weak supervision \cite{ren2020instance}, active-learning \cite{aghdam2019active} and self-supervision~\cite{gidaris2021obow} with task fine-tuning.

We consider here the extreme case of localizing objects in images without any annotation. Early works investigate regions proposals based on saliency \cite{zitnick2014edge} or intra-image similarity \cite{uijlings2013selective}, \ie, only between patches within the considered image (not across the image collection). However, these proposals have low precision and are produced in large quantities only to reduce the search space in other tasks, such as supervised~\cite{girshick2014cvpr_rcnn,girshick2015ICCV_fast_rcnn} or weakly-supervised~\cite{bilen2016weakly,tang2018pcl} object detection. Often using region proposals as input, unsupervised object discovery leverages information from the entire image collection and explores inter-image similarities to localize objects in an unsupervised fashion, \eg, with probabilistic matching~\cite{Cho_2015_CVPR}, principal component analysis~\cite{Wei2019ddtplus}, optimization~\cite{Vo_2019_CVPR,Vo20rOSD} and ranking~\cite{vo2021largescale}. However, because of the quadratic complexity of region comparison among images, together with the high number of region proposals for a single image, these methods hardly scale to large datasets. Other approaches do not require annotations but exploit extra modalities, \eg, audio~\cite{afouras2021self} or LiDAR~\cite{Tian_2021_CVPR}.

We propose here a simple approach to localize objects in an image, that we then apply to \emph{unsupervised object discovery}.
Our localization method stays at the level of a single image, rather than exploring inter-image similarity, which makes it linear \wrt the number of images and thus highly scalable. 
For this, we leverage high-quality features obtained from a visual transformer pre-trained with DINO self-supervision \cite{caron2021emerging}. 
Concretely, we divide the image of interest into equal-sized patches and feed it to the DINO model. 
Instead of focusing on the \cls token, we propose to use the \emph{key} component of the last attention layer for computing the similarities between the different patches. 
In doing so, we are able to localize a part of an object by selecting the patch with the least number of similar patches, here called the \emph{seed}. 
The justification for this seed selection criterion is based on the empirical observation that patches of foreground objects are less correlated than patches corresponding to background.
We add to this initial seed other patches that are highly correlated to it and thus likely to be part of the same object, a process which we call \emph{seed expansion}.
Finally, we construct a binary object segmentation mask by computing the similarities of each image patch to the selected seed patches and infer the bounding box of an object as the box that tightly encloses the largest connected component in this mask that contains the initial seed.
In following this simple method, we not only outperform methods for region proposals but also those for single-object discovery.
Even more, by training an off-the-self class-agnostic object detector using our localized boxes as ground-truth boxes, we are able to derive a much more accurate object localization model that is actually able to detect multiple objects in an image. 
We call this task \emph{unsupervised class-agnostic object detection} (which may resort to self-supervision despite being called unsupervised). 
Finally, by using clustering techniques to group the localized objects into visual consistent classes, we are able to train class-aware object detectors without any human supervision, but using instead the predicted object locations and their cluster ids as ground-truth annotations.
We call this task \emph{unsupervised (class-aware) object detection}.
We show that the predictions of our unsupervised detection model for certain clusters correlate very well with labelled semantic classes in the dataset and reach for them detection results competitive to object detectors trained with weak supervision~\cite{bilen2016weakly,tang2018pcl}.

Our main contributions are as follows:
(1) we show how to extract relevant features from a self-supervised pre-trained vision transformer and use the patch correlations within an image to propose a simple single-object localization method with linear complexity w.r.t. to dataset size;
(2) we leverage it to train both class-agnostic and class-aware unsupervised object detectors able to accurately localize multiple object per image and, in the class-aware case, group them to semantically-coherent classes;
(3)~we outperform the state of the art in unsupervised object discovery with a significant margin.
\section{Related work}
\label{sec:related}

\paragraph{Object detection with limited supervision.}
Region proposal methods~\cite{alexe2012objectness, uijlings2013selective, zitnick2014edge} generate in an unsupervised way numerous class-agnostic bounding boxes with high recall but low precision, to speed-up sliding window search. From supervised pre-trained networks, objects can emerge by masking the input~\cite{bergamo2016self}, interpreting neurons~\cite{zhou2015object} or from saliency maps~\cite{selvaraju2017grad}. Weakly-supervised object detection (WSOD) uses image-level labels without bounding boxes~\cite{bilen2016weakly, tang2018pcl} to learn to detect objects. The different instances of WSOD (each with specific assumptions on the availability and amount of image-level and box-level annotations) are often addressed as semi-supervised learning~\cite{gao2019note, tang2021proposal} and leverage self-training~\cite{radosavovic2017data, jie2017deep}. Recent work replaces manual annotations with automatic supervision from a different modality, \eg, LiDAR~\cite{Tian_2021_CVPR} or audio~\cite{afouras2021self}. In contrast, we do not use any annotations or other modalities at any stage: we extract object candidates from the activations of a self-supervised pre-trained network, compute pseudo-labels and then train an object detector.  

\paragraph{Object discovery.} Given a collection of images, object discovery groups images depicting similar objects, and then localizes objects within these images. Early works~\cite{grauman2006unsupervised,Russell06ObjectDiscovery,Sivic05ObjectDiscovery,TangLewis2008uod,Weber2000ObjectDiscovery} focus mostly on the first task and to, a lesser extent, on localization~\cite{Russell06ObjectDiscovery,Sivic05ObjectDiscovery,Zhang2015iccv_aog,Zhu2012cvpr_bmcl}.
On the contrary, \cite{Cho_2015_CVPR,kim2009pagerank_uod,Vo_2019_CVPR,Vo20rOSD,vo2021largescale} shift focus on the second task and achieve good object localization on image collections in the wild. However, casting object discovery as the selection of recurring visual patterns across an image collection involves expensive computation and only~\cite{vo2021largescale} is able to scale to large datasets. 
Our work also discovers object locations but does not consider inter-image similarity. Instead, we rely on the power of self-supervised transformer features \cite{caron2021emerging} and only consider intra-image similarity. Consequently, our method can localize objects in a single image with little computation. Close to ours,~\cite{zhang2020object} is also able to localize objects from a single image by exploiting scale-invariant features. Finally, some works~\cite{burgess2019monet,engelcke2019genesis,greff2019multi,locatello2020object, monnier2021decomposition} on object discovery attempt to simultaneously learn an image representation and to decompose images into object masks. These works, however, are only evaluated on image collections of very simple geometric objects.

\paragraph{Transformers.} In this work, we leverage transformer representations to address object discovery. Self-attention layers have been previously integrated into 
CNNs~\cite{hu2018relation, wang2018non, carion2020detr}, yet transformers for vision are very recent ~\cite{ramachandran2019stand, cordonnier2020relationship, chen2020generative, dosovitskiy2020vit} and still in an incipient stage. Findings on training heuristics~\cite{touvron2020deit, zhai2021scaling} and architecture design~\cite{liu2021swin, touvron2021going, yuan2021tokens} are released at high pace. Early adaptations of transformers to different tasks (\eg, image classification~\cite{dosovitskiy2020vit}, retrieval~\cite{tran_retrieval}, object detection~\cite{carion2020detr, zhu2021deformable, liu2021swin} and semantic segmentation~\cite{liu2021swin, strudel2021segmenter, xie2021segformer}) have demonstrated their utility and potential for vision. Meanwhile, several works attempt to better understand this new family of models from various perspectives~\cite{caron2021emerging, naseer2021intriguing, tuli2021convolutional, bhojanapalli2021understanding, minderer2021revisiting}.
Interestingly, transformers have been shown to be less biased towards textures than CNNs~\cite{tuli2021convolutional, naseer2021intriguing}, hinting that their features encapsulate more object-aware representations. These findings motivate us to study manners of localizing objects from transformer features.

\paragraph{Self-supervised learning (SSL)} is a powerful training scheme to learn useful representations without human annotations. It does so via a pretext learning task for which the supervision signal comes from the data itself~\cite{noroozi2016unsupervised, gidaris2018unsupervised, zhang2016colorful}. SSL pre-trained networks have been shown to outperform ImageNet pre-trained networks on several computer vision tasks, in particular object detection~\cite{gidaris2020learning, he2020momentum, caron2020unsupervised, grill2020bootstrap, gidaris2021obow}. For transformers, SSL methods also work well~\cite{caron2021emerging, xie2021self}, bringing a few interesting side-effects. In particular, DINO~\cite{caron2021emerging} feature activations appear to contain explicit information about the semantic segmentation of objects in an image.
In the same spirit, we extract another kind of transformer features to build our object localization.
\section{Proposed approach}
\label{sec:method}

Our method exploits image representations extracted by a vision transformer. In this section, we first recall how such representations are obtained, then present our method.

\subsection{Transformers for Vision}

\paragraph{Input.}
Vision transformers operate on a sequence of patches of fixed size $P {\times} P$. For a color image $\vI$ of spatial size $H {\times} W$, we have $N = H W / P^2$ patches of size $3 P^2$ (we assume for simplicity that $H$ and $W$ are multiples of $P$). Each patch is first embedded in a $d$-dimensional latent space via a trained linear projection layer. An additional, learned vector called the ``class token'', \cls, is adjoined to the patch embeddings,
yielding a transformer input in $\real^{(N+1) \times d}$.

\paragraph{Self-attention.} Transformers consist of a sequence of multi-head self-attention layers 
and multi-layer perceptrons (MLPs)~\cite{2017attention,dosovitskiy2020vit}. 
Three different learned linear transformations are applied to an input $\vX \in \real^{(N+1)\times d}$ of a self-attention layer to produce a query $\vQ$, a key $\vK$ and a value $\vV$, all in $\real^{(N + 1)\times d}$. The output of the self-attention layer is $\vY =  \text{softmax}\left( d^{-1/2} \, \vQ \vK\tran \right) \vV \in \real^{(N+1) \times d}$, where {\rm softmax} is applied row-wise. 
For simplicity, we describe here the case of a single-head attention layer, but attention layers usually contain multiple heads. In this work, we concatenate the keys (or queries, or values) from all heads in the last self-attention layer to obtain our feature representations.

\paragraph{Features for object localization.} 
We use transformers trained in a self-supervised manner using DINO \cite{caron2021emerging}.
Caron et al.\ \cite{caron2021emerging} show that sensible object segmentations can be obtained from the self-attention of the \cls query produced by the last attention layer. We adapt this strategy in \autoref{sec:exp} to perform object localization, providing a baseline (`DINO-seg') that produces fair results. 
However, we found that its does not fully exploit the potential of the self-supervised transformer features.
We propose a novel and 
effective strategy for localizing objects using another way to extract and use features.
Our method, called \ours, is constructed by computing similarities between patches of a single image, using this time patch keys $\vk_p \,{\in}\, \real^d$, $p=1, \ldots, N$, extracted at the last layer of a transformer. 

\subsection{Finding objects with \ours} \label{sec:finding_objects}

\begin{figure*}[t]
\centering
    \includegraphics[height=25mm]{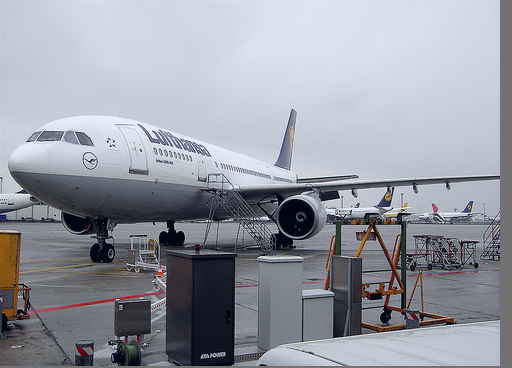}
    \includegraphics[height=25mm]{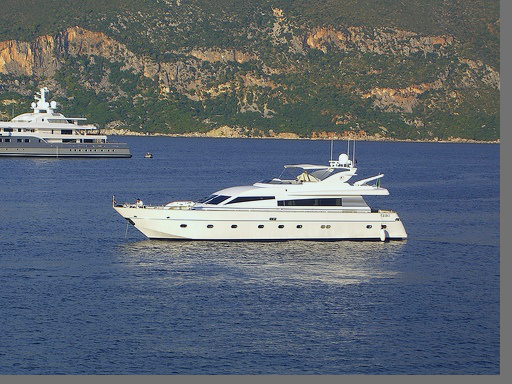}
    \includegraphics[height=25mm]{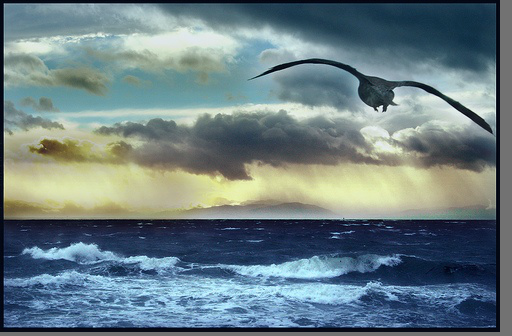}
    \includegraphics[height=25mm]{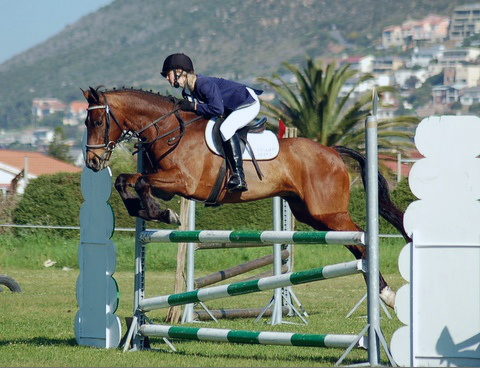}\\
    \includegraphics[height=25mm]{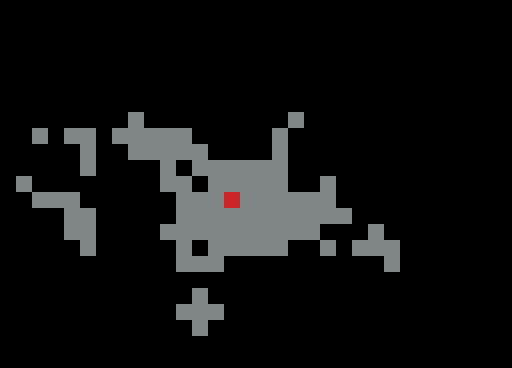}
    \includegraphics[height=25mm]{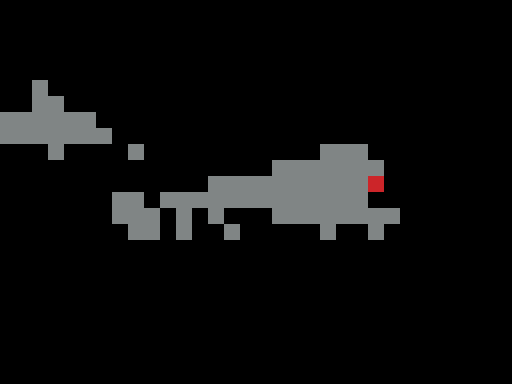}
    \includegraphics[height=25mm]{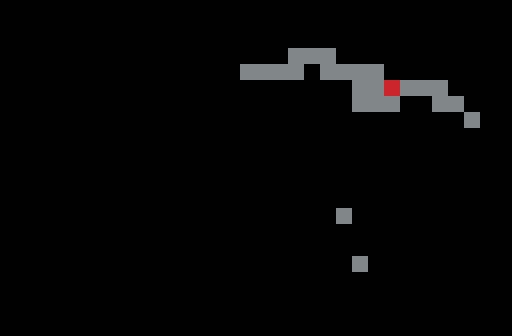}
    \includegraphics[height=25mm]{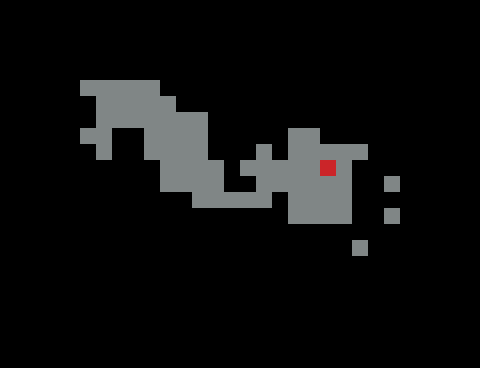}\\
    \includegraphics[height=25mm]{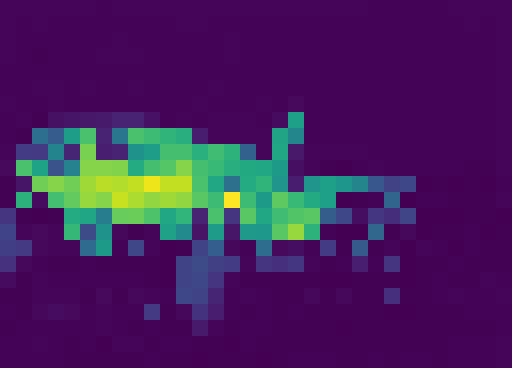}
    \includegraphics[height=25mm]{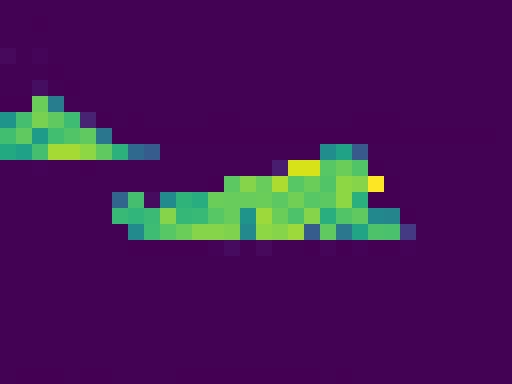}
    \includegraphics[height=25mm]{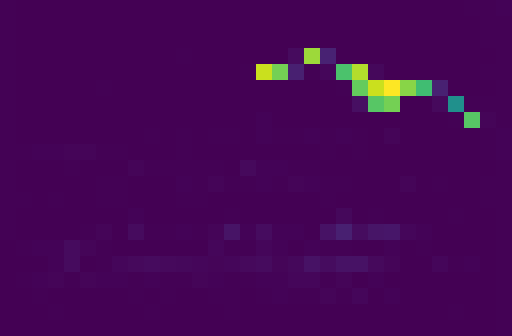}
    \includegraphics[height=25mm]{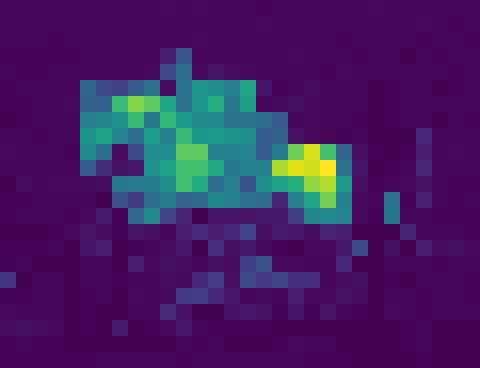}\\
\caption{\small
\textbf{Initial seed, patch similarities and patch degrees.} Top: images from Pascal VOC2007. Middle: initial seed $p^*$ (in red) and patches similar to $p^*$ (in grey), \ie, such that $\vf_p^{\smash{\tran}} \vf_q \geq 0$ hence $a_{p^*q}=1$. Bottom: map of inverse degrees $1/d_p$ of all patches $p$ (yellow to blue, for low to high degrees). The initial seed $p^*$ is the patch with the lowest degree. Figure is best viewed in color.}
\label{fig:similarity_degree}
\end{figure*}

Our method takes as input $d$-dimensional image features $\vF \in \real ^{N \times d}$ extracted from a single image via a neural network; $N$ denotes the spatial dimension (number of patches) of the image features~$\vF$, while $\vf_{p} \in \real ^{d}$ is the feature vector of the patch at spatial position $p \in \{1, \ldots, N\}$. We assume that there is at least one object in the image and \ours tries to localize one of them given the input features. To that end, it relies on a selection of patches that are likely to belong to an object. We call these patches ``seeds''. 

\paragraph{Initial seed selection.}
Our seed selection strategy is based on the assumptions that (a)~regions/patches within objects correlate more with each other than with background patches and vice versa, and (b)~an individual object covers less area than the background. Consequently, a patch with little correlation in the image has higher chances to belong to an object.

To compute the patch correlations, we rely on the distinctiveness of self-supervised transformer features, which is particularly noticeable when using transformer's keys.
We empirically observe that using these tranformer features as patch representation meets assumption~(a) in practice: patches in an object correlate positively with each other but negatively with patches in the background. Therefore, based on assumption (b), we select the first seed $p^*$ by picking the patch with the smallest number of positive correlations with other patches. 

Concretely, we build a patch similarity graph $\mathcal{G}$ per image, represented by the binary symmetric adjacency matrix $\vA \,{=}\, (a_{pq})_{1 \leq p, q\leq N} \in \{0, 1\}^{N \times N}$ such that
\begin{align}
\label{eq:sim}
a_{pq} = 
\left\{
    \begin{array}{ll}
    1 & \text{if } \vf_p^{\smash{\tran}} \vf_q \geq 0, \\
    0 & \text{otherwise}.
    \end{array}
\right.
\end{align}
In other words, two nodes $p,q$ are connected by an undirected edge if their features $\vf_p, \vf_q$ are positively correlated. Then, we select the initial seed $p^*$ as a patch with the lowest degree~$d_p$:
\begin{align}
\label{eq:deg}
    p^* = \argmin_{p \in \{1, \ldots, N\}} d_p \text{~~~where~~~} d_p = \sum_{q=1}^{N} a_{pq}.
\end{align}
We show in \autoref{fig:similarity_degree} examples of seeds $p^*$ selected in four different images. A representation of the degree map for each of these images is also presented. We remark that the patches with lowest degrees are the most likely to fall in an object. Finally, we also observe in this figure that the few patches that correlate positively with $p^*$ are also likely to belong to an object.

\begin{figure*}
\centering
\includegraphics[height=28mm]{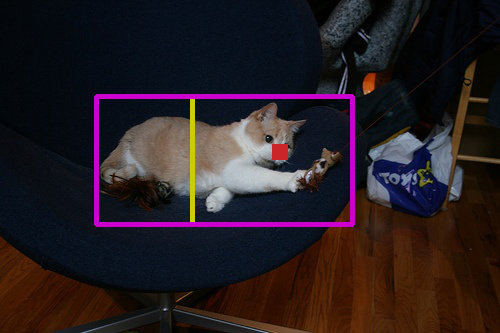}
\includegraphics[height=28mm]{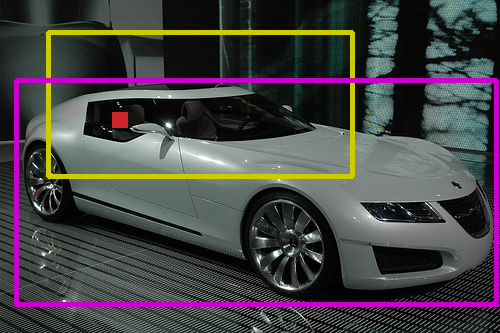}
\includegraphics[height=28mm]{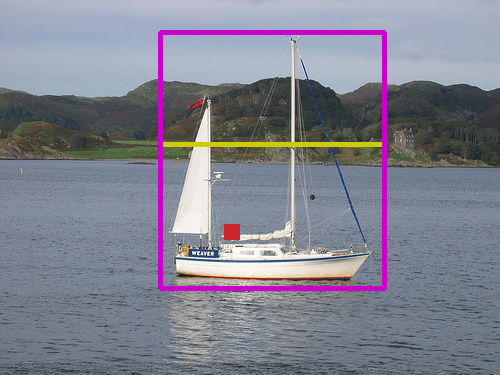}
\includegraphics[height=28mm]{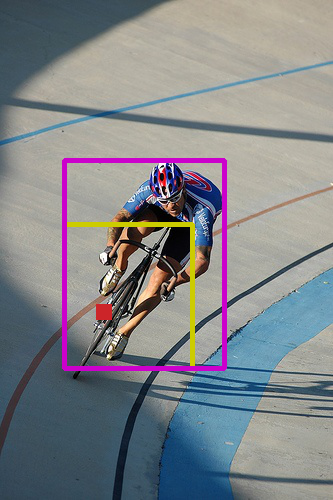}
\caption{\small
\textbf{Object localizations on VOC07}. The red square represents the seed $p^*$, the yellow box is the box obtained using only the seed $p^*$, and the purple box is the box obtained using all the seeds $\mathcal{S}$.
}
\label{fig:similars}
\end{figure*} 

\paragraph{Seed expansion.} Once the initial seed is selected, the second step consists in selecting patches correlated with the seed that are also likely to fall in the object. Again, we achieve this step relying on the empirical observations that pixels within an object tend to be positively correlated and to have a small degree in $\mathcal{G}$. We select the next best seeds after $p^*$ as the pixels that are positively correlated with $\vf_{p^*}$: $\mathcal{S} = \{q \mid q \in \mathcal{D}_k \text{ and } \vf_q^{\tran} \vf_{p^*} \geq 0\}$ within $\mathcal{D}_k$, the $k$ patches with the lowest degree.
 (In case of patches with equal degrees, we break ties arbitrarily to ensure that $\vert \mathcal{D}_k \vert=k$.)
Note that $p^* \in \mathcal{D}_k$ and a typical value for $k$ is $100$.

\paragraph{Box extraction.} The last step consists in computing a mask $\vm~\in~\{0, 1\}^N$ by comparing the seed features in $\mathcal{S}$ with all the image features. The $q^{\rm th}$ entry of the mask $\vm$ satisfies
\begin{align}
\label{eq:box-extraction}
m_q = 
\left\{
    \begin{array}{ll} 
    1 & \text{if } \sum_{s \in \mathcal{S} } \vf_q^{\tran} \vf_{s} \geq 0, \\
    0 & \text{otherwise}.
    \end{array}
\right.
\end{align}
In other words, a patch $q$ is considered as part of an object if, on average, its feature $\vf_q$ positively correlates with the features of the patches in $\mathcal{S}$. To remove the last spurious correlated patches, we finally select the connected component in $\vm$ that contains the initial seed and use the bounding box of this component as the detected object. An illustration of the detected boxes before and after seed expansion is provided in \autoref{fig:similars}.

\subsection{Towards unsupervised object detection}

We exploit the accurate single-object localization of \ours for training object detection models without any human supervision.
Starting from a set of unlabeled images, each one assumed to contain at least one prominent object, we extract one bounding box per image using \ours. Then, we train off-the-shelf object detectors using these pseudo-annotated boxes. We explore two scenarios: class-agnostic and (pseudo) class-aware training of object detectors.

\paragraph{Class-agnostic detection (\cad).}
A class-agnostic detection model localizes salient objects in an image without predicting nor caring about their semantic category.
We train such a detector by assigning the same ``foreground'' category to all the boxes produced by \ours, which we call ``pseudo-boxes'' afterwards, as they are obtained with no supervision.
Unlike \ours, the trained detector can localize multiple objects per image, even if it was trained on a dataset containing only one pseudo-box annotation per image. The experiments confirm that the trained detector can output multiple detections and the quantitative results (\autoref{tab:training}) show that this trained detector is in fact  even better than \ours in terms of localization accuracy.

\paragraph{Class-aware detection (\od).}
We now consider a typical detector that both localizes objects and recognizes their semantic category.
To train such a detector, apart from \ours's pseudo-boxes, we also need a class label for each of these boxes.
In order to remain fully-unsupervised,
we discover visually-consistent object categories using K-means clustering.
For each image, we crop the object detected by \ours, resize the cropped image to $224 \times 224$, feed this image in the DINO pre-trained transformer, and extract the \cls token at the last layer. The set of \cls tokens are clustered using K-means and the cluster index is used as a pseudo-label for training the detector.
At evaluation time, we match these pseudo-labels to the ground-truth class labels using the Hungarian algorithm \cite{kuhn1955hungarian}, which give names to pseudo-labels.
\section{Experiments}
\label{sec:exp}

We explore in this section three variants of the object
localization problem, in order of increasing complexity: (1)
localizing one salient object in each image (single-object discovery) in~\S\ref{sec:results_localization}, (2) using the corresponding
bounding boxes as ground-truth to train a binary classifier for
foreground object detection (unsupervised class-agnostic object detection), and (3) using clustering to capture an
unsupervised notion of object categories, and detect the corresponding
instances (unsupervised object detection). Both are discussed in~\S\ref{sec:results_detection}. None of the building blocks of this pipeline uses any
annotation, just a large number of unlabelled images to sequentially
train, in a self-supervised way, the DINO transformer, the class
agnostic foreground/background classifier, and finally the classifier
using the cluster identifier as labels.
Also, we provide more qualitative results in supplementary.

\subsection{Experimental setup}
\paragraph{Backbone networks.} 
Unless otherwise specified, we use the ViT-S model introduced in \cite{caron2021emerging}, which follows the architecture of DEiT-S \cite{touvron2020deit}. It is trained using DINO~\cite{caron2021emerging}, with a patch size of $P=16$ and the keys $\vK$ (without the entry corresponding to the \cls token) of the last layer as input features $\vF$, with which we achieve the best results. Results obtained alternatively with the attention, the queries and values are presented and discussed in the supplementary material.
For comparison, we also present results using the base version of ViT (ViT-B), ViT-S with a patch size of $P=8$, as well as with features of the last convolutional layer of a dilated ResNet-50 \cite{He2016cvpr_resnet} and of a VGG16 \cite{Symonian2014verydeep} pre-trained either following DINO, or in a supervised fashion on Imagenet \cite{imagenet_cvpr09}.

\paragraph{Datasets.} 
We evaluate the performance of our approach on the three variants of object localization 
on VOC07~\cite{pascal-voc-2007} trainval+test,
VOC12~\cite{pascal-voc-2012} trainval and COCO\_20K~\cite{Lin2014cocodataset,Vo20rOSD}. VOC07 and VOC12 are commonly used benchmarks for object detection~\cite{girshick2014cvpr_rcnn,girshick2015ICCV_fast_rcnn}. COCO\_20k is a subset of the COCO2014 trainval dataset~\cite{Lin2014cocodataset}, consisting of 19817 randomly chosen images, used as a benchmark in~\cite{Vo20rOSD}. When evaluating results on the unsupervised object discovery task, we follow a common practice and evaluate scores on the trainval set of the different datasets. Such an evaluation is possible as the task is fully unsupervised. We follow the same principle for the unsupervised class-agnostic task: we generate boxes on VOC07 trainval, VOC12 trainval and COCO\_20k, use them to train a class-agnostic detector, and then evaluate again on these datasets (against ground-truth boxes this time). For unsupervised class-aware object detection, we generate boxes and train the detector on VOC07 trainval and/or VOC12 trainval, but evaluate the detector on the VOC07 test set to facilitate comparisons to weakly-supervised object detection methods.
Note that for unsupervised object discovery, some previous works~\cite{Vo_2019_CVPR,Vo20rOSD,vo2021largescale,Wei2019ddtplus} evaluate on subsets of VOC07 trainval and VOC12 trainval. For completeness, we present the object discovery performance of our method on these reduced datasets in the supplemental material.

\subsection{Application to unsupervised object discovery} \label{sec:results_localization}

\begin{table}[t]
    \footnotesize
    \centering
    \begin{tabular}{lcccc}
    	\toprule
    	Method &  VOC07\_trainval & VOC12\_trainval & COCO\_20k \\
    	\midrule
    	Selective Search~\cite{uijlings2013selective} & 18.8 & 20.9 & 16.0  \\
    	EdgeBoxes~\cite{zitnick2014edge} & 31.1 & 31.6 & 28.8 \\
    	Kim \etal \cite{kim2009pagerank_uod} & 43.9 & 46.4 & 35.1  \\
    	Zhang \etal \cite{zhang2020object} & 46.2 & 50.5 & 34.8  \\
    	DDT+ \cite{Wei2019ddtplus} & 50.2 & 53.1 & 38.2 \\
    	rOSD \cite{Vo20rOSD} &  54.5 & 55.3 & 48.5  \\
    	LOD \cite{vo2021largescale} & 53.6 & 55.1 & 48.5  \\
    	\midrule
    	DINO-seg (w. ViT-S/16) & 45.8 & 46.2 & 42.1 \\
    	\ours (ours) & \bf 61.9  & \bf 64.0 & \bf 50.7\\
    	\midrule
    	\midrule
    	rOSD \cite{Vo20rOSD} + \cad & 58.3 & 62.3 & 53.0 \\
    	LOD \cite{vo2021largescale} + \cad & 56.3 & 61.6 & 52.7 \\
    	\ours (ours) + \cad  & \bf 65.7 & \bf 70.4 & \bf 57.5 \\
    	\bottomrule
    \end{tabular}
    \vspace{5pt}
    \caption{\small \textbf{Single-object discovery.} CorLoc performance on VOC07 trainval, VOC12 trainval and COCO\_20k. We compare \ours to state-of-the-art object discovery methods~\cite{kim2009pagerank_uod,Vo20rOSD,vo2021largescale,Wei2019ddtplus,zhang2020object}, as well as to two object proposal methods \cite{uijlings2013selective, zitnick2014edge}. We also compare to the segmentation method proposed in DINO \cite{caron2021emerging}, denoted by DINO-seg. Additionally, we train a class-agnostic dectector (+ \cad) using as ground-truth either our pseudo-boxes or the boxes of rOSD \cite{Vo20rOSD} or LOD \cite{vo2021largescale}.}
    \label{tab:training}
\end{table}

Similar to methods for unsupervised single-object discovery, \ours produces one box for each image. It therefore can be directly evaluated for this task. Following \cite{Cho_2015_CVPR,Vo_2019_CVPR,Vo20rOSD,vo2021largescale,zhang2020object}, we use the \textit{Correct Localization} (CorLoc) metric, \ie, the percentage of correct boxes, where a predicted box is considered correct if it has an \textit{intersection over union} (IoU) score superior to $0.5$ with one of the labeled object bounding boxes.

\paragraph{Comparison to prior work.} 
In \autoref{tab:training}, we present the CorLoc of our method, in comparison to state-of-the-art object discovery methods~\cite{kim2009pagerank_uod,Vo20rOSD,vo2021largescale, Wei2019ddtplus,zhang2020object} and region proposals \cite{uijlings2013selective, zitnick2014edge}.

Despite its simplicity, we see that {\ours} outperforms the other methods by large margins. We also compare against an adapted version of the segmentation method proposed in \cite{caron2021emerging}. 
Concretely, we extract the self-attention of the \cls query at the last layer of the transformer, create a binary mask where the $0.6\,N$ largest entries of this self-attention are set to $1$, retrieve the largest spatially-connected component from this binary mask, and use the bounding box of this component as the detected object. 
This method returns one box per self-attention head and we report results obtained with the best performing head over the entire dataset, noted as DINO-seg. \ours improves over DINO-seg by 8 to 17  of CorLoc points, demonstrating the efficacy of our approach for object localization based on self-supervised pre-trained transformer features. 

Finally, we also evaluate our unsupervised class-agnostic detector (denoted by `+ \cad') for single-object discovery. To this end, we return for each image the box that the detector assigns the highest score. It can be seen that training a class-agnostic detector on \ours's outputs further improves the performance by 4 to 7 CorLoc points.
In total, our method surpasses the prior state of the art by at least $10$ CorLoc points on each evaluated dataset.

\paragraph{Impact of the backbone architecture.} 
\autoref{tab:backbone} studies the effect of the backbone on \ours. We see that transformer representations are better suited for our method (best results with ViT-S/16). In contrast, our performance using the DINO-pre-trained ResNet-50 is significantly lower. It indicates that the performance of our method is not only due to the contributions of self-supervision but also to the property and quality of the specific features we extract. 

\begin{table}[t]
    \centering
    \footnotesize
    \begin{tabular}{lccccc}
    	\toprule
    	Backbone & pre-training & VOC07\_trainval & VOC12\_trainval & COCO$\_$20k \\
    	\midrule
    	VGG16 & supervised & 42.0 & 47.2 & 30.2\\
    	ResNet50 & supervised & 33.5 & 39.1 & 25.5\\
    	\midrule
    	ResNet50 & DINO & 36.8 & 42.7 & 26.5\\
    	ViT-S/8 & DINO & 55.5 & 57.0 & 49.5 \\
    	ViT-S/16 & DINO & \bf 61.9 & \bf 64.0 & \bf 50.7 \\
    	ViT-B/16 & DINO & 60.1 & 63.3 & 50.0 \\
    	\bottomrule
    \end{tabular}
    \vspace{5pt}
    \caption{\small \textbf{Impact of the backbone.} We evaluate {\ours} on features originating from different backbones: ViT \cite{dosovitskiy2020vit} small (ViT-S) and base (ViT-B) with patch size $P\,{=}\,8 ~\text{or}~ 16$, ResNet50 \cite{He2016cvpr_resnet} pre-trained following DINO \cite{caron2021emerging}, and VGG16 \cite{Symonian2014verydeep} and ResNet50 trained in a fully-supervised fashion on Imagenet \cite{imagenet_cvpr09}.}
    \label{tab:backbone}
    \vspace{-5pt}
\end{table}

\begin{table}[t]
    \Huge
    \centering
    \resizebox{\textwidth}{!}{%
    \begin{tabular}{l|c|cccccccccccccccccccc|c}
    	\toprule
    	Method & Supervis. & aero & bike & bird & boat & bottle & bus & car & cat & chair & cow & table & dog & horse & mbike & person & plant & sheep & sofa & train & tv & mean\\
    	\midrule
    	WSDDN~\cite{bilen2016weakly} & weak & 39.4 & 50.1 & 31.5 & 16.3 & 12.6 & 64.5 & 42.8 & 42.6 & 10.1 & 35.7 & 24.9 & 38.2 & 34.4 & 55.6 & 9.4 & 14.7 & 30.2 & 40.7 & 54.7 & 46.9 & 34.8 \\
     	PCL~\cite{tang2018pcl} & weak & 54.4 & 9.0 & 39.3 & 19.2 & 15.7 & 62.9 & 64.4 & 30.0 & 25.1 & 52.5 & 44.4 & 19.6 & 39.3 & 67.7 & 17.8 & 22.9 & 46.6 & 57.5 & 58.6 & 63.0 & 43.5\\
     	\midrule
     	\midrule
     	rOSD~\cite{Vo20rOSD} + \od & none & 38.8 & 44.7 & 25.2 & 15.8 & 0.0 & 52.9 & 45.4 & 38.9 & 0.0 & 16.6 & 24.4 & 43.3 & 57.2 & 51.6 & 8.2 & 0.7 & 0.0 & 9.1 & 65.8 & 9.4 & 27.4 \\
     	\midrule
     	\ours  pseudo-boxes & none & 42.8 & 0.0 & 16.4 & 3.9 & 0.0 & 32.4 & 17.1 & 26.2 & 0.0 & 14.2 & 11.3 & 28.1 & 43.9 & 15.8 & 2.2 & 0.0 & 0.1 & 5.6 & 39.9 & 2.3 & 15.1\\
     	\ours + \od & none & 57.4 & 0.0 & 40.0 & 19.3 & 0.0 & 53.4 & 41.2 & 72.2 & 0.2 & 24.0 & 28.1 & 55.0 & 57.2 & 25.0 & 8.3 & 1.1 & 0.9 & 21.0 & 61.4 & 5.6 & 28.6\\
     	\ours + OD$^\dagger$ & none & 62.0 & 38.5 & 49.3 & 23.1 & 4.2 & 57.0 & 41.9 & 70.4 & 0.0 & 3.6 & 18.9 & 30.8 & 52.8 & 45.5 & 12.5 & 0.6 & 9.1 & 9.0 & 67.2 & 0.8 & \textbf{29.9}\\     	
    	\bottomrule
    \end{tabular}
    }
    \caption{\small \textbf{Object detection.} Results (in AP@0.5 \%) on VOC07 test.
   \ours\!+\,\od and rOSD\,\cite{Vo20rOSD}\,+\,\od are trained on VOC07 trainval. \ours + OD$^\dagger$ is trained on the union of VOC07 and VOC12 trainval sets. 
    }
    \label{tab:training-class-ap}
\end{table}

\begin{table}[t]
    \centering
    \footnotesize
    \begin{tabular}{l@{~~~}cccc}
    	\toprule
    	& \multicolumn{2}{c}{VOC07} & VOC12 & COCO20k \\
    	Training set (when applicable) & \multicolumn{2}{c}{trainval} & trainval  & trainval \\
    	Evaluation set & trainval & test & trainval & trainval \\
    	\midrule 
        \midrule 
    	EdgeBoxes~\cite{zitnick2014edge} &  3.6 & 4.4 & 4.8 & 1.8 \\
        Selective Search~\cite{uijlings2013selective} & 2.9 & 3.6 & 4.2 & 1.6 \\
    	\midrule
    	rOSD \cite{Vo20rOSD} + \cad & 24.2 & 25.2 & 29.0 & 8.4\\
    	LOD \cite{vo2021largescale} + \cad & 22.7 & 23.7 & 28.4 & 8.8\\
    	\ours + \cad & \textbf{29.0} & \textbf{29.0} & \textbf{33.5} & \textbf{9.9} \\
    	\bottomrule
    \end{tabular}
    \vspace{3pt}
    \caption{\small\textbf{Class-agnostic unsupervised object detection results} (in AP@0.5 $\%$).
    Trainings, corresponding to `\emph{method} + \cad', are performed on the bare images and rely only on the fully-unsupervised methods rOSD \cite{Vo20rOSD}, LOD \cite{vo2021largescale} and \ours (ours). Evaluation of unsupervised object detection may thus be performed on the same images as those used for unsupervised training (without manual annotations). The classic methods EdgeBoxes~\cite{zitnick2014edge} and Selective Search~\cite{uijlings2013selective} do not involve any training.
    }
    \label{tab:training-map-short}
    \vspace{-5pt}
\end{table}

\subsection{Unsupervised object detection} \label{sec:results_detection}

Here we explore the application of \ours in unsupervised object detection.
To that end, we use \ours's pseudo-boxes to train a Faster R-CNN model \cite{shaoqing2015faster} on the datasets. We measure detection performance using the \textit{Average Precision at IoU 0.5} metric (AP@0.5), which is commonly used in the PASCAL detection benchmark. As Faster R-CNN backbone, we use a ResNet50 pre-trained with DINO self-supervision, thus making our training pipeline fully-unsupervised.
We trained the Faster R-CNN models using the detectron2 \cite{wu2019detectron2} implementation (more details in the supplementary material).

\paragraph{Pseudo-labels.}  
To generate pseudo-labels for the class-aware detectors, we apply K-means clustering on DINO-ViT-S tokens using as many clusters as the number of different classes in the dataset.
Since the cluster-based pseudo-labels are ``anonymous'', to evaluate the detection results we must map the clusters to the ground-truth classes.
Following prior work in image clustering~\cite{asano2019self,bautista2016cliquecnn,ji2019invariant}, we use Hungarian matching~\cite{kuhn1955hungarian} for that.
We stress that this matching is only for reporting evaluation results; we do not use any human labels during training.

\paragraph{Unsupervised class-aware detection.} 
\autoref{tab:training-class-ap} provides results of unsupervised class-aware object detectors trained with \ours (entry `\ours + \od'). 
We are not aware of any prior work that addresses unsupervised object detection on real-world images of complex scenes, as those in PASCAL, that does not use extra modalities. 
We could not compare to \cite{triantafyllos2020self, Tian_2021_CVPR} as we focus on image-only benchmarks.

We see that, although fully-unsupervised, our method learns to accurately detect several object classes.
For example, detection performance for classes ``aeroplane'', ``bus'', ``dog'', ``horse'' and ``train'' is more than $50.0\%$, and for ``cat'' it reaches $72.2\%$.
Even more so, for some classes our method achieves better AP than the weakly-supervised methods 
WSDDN~\cite{bilen2016weakly} and PCL~\cite{tang2018pcl}, which require image-wise human labels.
Although the results are not entirely comparable due to backbone differences between our method and the weakly-supervised ones (self-supervised ResNet50 vs. supervised VGG16),
they still demonstrate the efficacy of our method in unsupervised object detection, which is an extremely hard and ill-posed task.

We also evaluate the AP of our pseudo-boxes (with their assigned cluster id as pseudo-labels) when generated for VOC07 test (entry `\ours pseudo-boxes'). 
Evidently, training the detector on pseudo-boxes leads to a significantly higher AP than the initial pseudo-boxes.

Finally, switching our pseudo-boxes with those of rOSD~\cite{Vo20rOSD} for the detector training (adding pseudo-labels to rOSD pseudo-boxes by clustering DINO features in exactly the same way as in our method) leads to performance degradation (entry `rOSD + \od').

\paragraph{Unsupervised class-agnostic detection.} 
In \autoref{tab:training-map-short}, we report class-agnostic detection results obtained using pseudo-boxes from our method (`\ours + \cad') as well as from rOSD~\cite{Vo20rOSD} (`rOSD + \cad') and LOD~\cite{vo2021largescale}  (`LOD + \cad').
As we see, our method leads to a significantly better detection performance.
We also report detection results using the Selective Search~\cite{uijlings2013selective} and EdgeBox\cite{zitnick2014edge} proposal algorithms, which perform worse than our method.

\subsection{Limitations and future work}

Despite the good performance of \ours, it exhibits some limitations.

\ours, as it stands, can separate same-class instances that do not overlap (as it only keeps the connected component of the initial seed to create a box), but it is not designed to separate instances when overlapping.
This is actually a challenging problem, related to the difference between supervised semantic~\cite{Long2015fcn} and instance~\cite{He2017ICCV_mask_rcnn} segmentation methods, which, as far as we know, is an open problem in the absence of any supervision. A potential lead could be to use a matching algorithm such as Probabilistic Hough Matching to separate instances within image regions found in multiple images.

Another issue is when an object covers most of the image. 
It violates our second assumption for the initial seed selection (expressed in \autoref{sec:finding_objects}) that an individual object covers less area than the background, thus possibly causing the seed to fall in the background instead of a foreground object.
Ideally, we would like to filter out such failure cases, e.g., by using the attention maps of the \cls token. We leave this as future work.

\section{Conclusion}
We have presented \ours, a simple, yet effective method for localizing objects in images without any labels, 
by leveraging self-supervised pre-trained transformer features~\cite{caron2021emerging}.
Despite its simplicity, 
{\ours} outperforms state-of-the-art methods in object discovery by large margins. Having high precision, the boxes found by {\ours} can be used as pseudo ground truth for training a class-agnostic detector which further improves the object discovery performance. 
\ours boxes can also be used to train an unsupervised object detector that yields competitive results compared to weakly-supervised counterparts for several classes.

Future work will be dedicated to investigate other applications of \ours boxes, \eg, high-quality region proposals for object detection tasks, and the power of self-supervised transformer features for unsupervised object segmentation. 
 
\section{Acknowledgments and Disclosure of Funding}
This work was supported in part by the Inria/NYU collaboration, the Louis Vuitton/ENS chair on artificial intelligence and the French government under management of Agence Nationale de la Recherche as part of the ``Investissements d’avenir'' program, reference ANR19-P3IA-0001 (PRAIRIE 3IA Institute). Huy V. Vo was supported in part by a Valeo/Prairie CIFRE PhD Fellowship.
 
\clearpage
\bibliography{main} 
\bibliographystyle{plain}

\clearpage
\appendix

\section{Ablation Study}

\subsection{Which transformer features to choose?}

As explained in \autoref{sec:finding_objects} of the main paper, we chose to use the keys $\vk_p$ of the last attention layer as patch features $\vf_p$ in \ours.
As we will see here, this choice provides the best localization performance among other alternatives. 
Specifically, in the first section of \autoref{tab:paramk}, we report the performance of \ours when using as patch features $\vf_p$ either the keys~$\vk_p$, the queries~$\vq_p$, or the values~$\vv_p$ of the attention layer. We see that, when using the queries~$\vq_p$ or the values~$\vv_p$, \ours's performance deteriorates by at least 11 CorLoc points compared to using the keys~$\vk_p$.

Another way to measure the similarity between two patches in a transformer architecture is to use the scalar product between the queries and the keys. We thus test substituting 
\begin{align}
\tilde{a}_{pq} = 
\left\{
    \begin{array}{ll}
    1 & \text{if } \vq_p^{\smash{\tran}} \vk_q + \vk_p^{\smash{\tran}} \vq_q \geq 0, \\
    0 & \text{otherwise},
    \end{array}
\right.
\label{eq:sim_qk}
\end{align}
for ${a}_{pq}$ in Eq.\,(\ref{eq:sim}) in the main paper, when selecting the first, initial seed. Note that this choice of $\tilde{a}_{pq}$ ensures the symmetry of the adjacency matrix. We test this new choice of similarity matrix when using the queries, keys or values in the seed expansion step, i.e., in $\mathcal{S}$, and in the box extraction steps, i.e., in $\vm$ as defined in Eq.\,(\ref{eq:box-extraction}) in the main paper. 

Finally, we also test another alternative by changing the definition of $\mathcal{S}$ to $\tilde{\mathcal{S}} = \{q \mid q \in {\mathcal{D}}_k \text{ and } \vq_q^{{\tran}} \vk_{p^*} + \vk_q^{{\tran}} \vq_{p^*} \geq 0\}$ and changing the definition of $m_q$ in Eq.\,(\ref{eq:box-extraction}) to
\begin{align}
\tilde{m}_q = 
\left\{
    \begin{array}{ll} 
    1 & \text{if } \sum_{s \in \mathcal{S} } \left( \vk_q^{\tran} \vq_{s} + \vq_q^{\tran} \vk_{s} \right) \geq 0, \\
    0 & \text{otherwise}.
    \end{array}
\right.
\end{align}
Results in \autoref{tab:paramk} show that all these alternatives using queries and keys yield results that are not as good as when using the keys as patch features.

\begin{table}[t]
\centering
\small    
\begin{tabular}{llll}
\toprule
	Seed selection 
	    & Expansion \& Box extrac.
	    & $k$ 
	    & CorLoc\\
\midrule
	$a_{pq}$ with $\vf_{p,q}$ = $\vq_{p,q}$ in Eq.~(1)
	    & $\vf_p = \vq_p$ in $\mathcal{S}$ and $m_q$
	    & 100 
	    & 30.8 \\
	$a_{pq}$ with $\vf_{p,q}$ = $\vv_{p,q}$ in Eq.~(1)
	    & $\vf_p = \vv_p$ in $\mathcal{S}$ and $m_q$
	    & 100 
	    & 50.5\\
	$a_{pq}$ with $\vf_{p,q}$ = $\vk_{p,q}$ in Eq.~(1)
	    & $\vf_p$ = $\vk_p$ in $\mathcal{S}$ and $m_q$
	    & 100 
	    & 61.9 \\
\midrule
    $\tilde{a}_{pq}$ defined in \ref{eq:sim_qk}
	    & $\vf_p$ = $\vq_p$ in $\mathcal{S}$ and $m_q$
	    & 100 
	    & 30.8 \\
    $\tilde{a}_{pq}$ defined in \ref{eq:sim_qk}
	    & $\vf_p$ = $\vv_p$ in $\mathcal{S}$ and $m_q$
	    & 100 
	    & 29.9 \\
	$\tilde{a}_{pq}$ defined in \ref{eq:sim_qk}
	    & $\vf_p$ = $\vk_p$ in $\mathcal{S}$ and $m_q$
	    & 100 
	    & 30.7 \\
	$\tilde{a}_{pq}$ defined in \ref{eq:sim_qk}
	    & using $\tilde{\mathcal{S}}$ and $\tilde{m_q}$
	    & 100 
	    & 30.8 \\ 
\midrule
    $a_{pq}$ with $\vf_{p,q}$ = $\vk_{p,q}$ in Eq.~(1)
	    & $\vf_p$ = $\vk_p$ in $\mathcal{S}$ and $m_q$
	    & 1 
	    & 38.3 \\
    $a_{pq}$ with $\vf_{p,q}$ = $\vk_{p,q}$ in Eq.~(1)
	    & $\vf_p$ = $\vk_p$ in $\mathcal{S}$ and $m_q$
	    & 50
	    & 58.8 \\
	$a_{pq}$ with $\vf_{p,q}$ = $\vk_{p,q}$ in Eq.~(1)
	    & $\vf_p$ = $\vk_p$ in $\mathcal{S}$ and $m_q$
	    & 150 
	    & 61.8 \\
	$a_{pq}$ with $\vf_{p,q}$ = $\vk_{p,q}$ in Eq.~(1)
	    & $\vf_p$ = $\vk_p$ in $\mathcal{S}$ and $m_q$
	    & 200 
	    & 61.2 \\
\bottomrule
\end{tabular}
\vspace{5pt}
\caption{\small \textbf{Ablation study.} CorLoc performance on VOC2007 for different choices of transformer features in the seed selection, expansion and box extraction steps, as well as influence on the results of the parameter $k$ (maximum number of patches with the lowest degree, in $\mathcal{D}_k$, for seed expansion).}
\label{tab:paramk}
\end{table}

\subsection{Importance of the seed expansion step}
\label{sec:expansion}
We analyse here the importance of the seed expansion step that is controlled by $k$. The seed expansion step allows us to enlarge the region of interest so as to include all the parts of an object and not only the part localized from the first, initial seed.

\autoref{tab:paramk} presents the impact of the parameter~$k$, which corresponds to the maximum number of patches that can be used to construct the mask $\vm$. We notice that, without seed expansion (i.e., $k=1$), there is a drastic drop in the localization performance.
The performance then improves when increasing $k$ to $100$-$150$ with a slight decrease at $200$.

Visualizations of results with $k=1$ and $k=100$ are presented in the Figure 3 of the main paper and \autoref{fig:seed_expansion} here. We see that the boxes in yellow obtained with $k=1$ are small and localized on probably what is the most discriminative part of the objects. Increasing $k$ permits us to increase the size of the box and localize the object better. We also present in \autoref{fig:seed_expansion_failure} cases of failures where the seed expansion step is either insufficient to localize the whole object or yields a box containing multiple objects.

\subsection{Analysis of DINO-seg}
In this section, we investigate alternative setups of the baseline DINO-seg which is based on the work of Caron \etal \cite{caron2021emerging}. They are presented in \autoref{tab:dinoseg}.

First, instead of using the best attention head over the entire dataset (as we did in the main paper), we evaluate the localization accuracy of DINO-seg for each one of the 6 available heads. We find out that one head in particular, namely head~4, captures objects well, whilst results with other heads are much lower. Due to its superior performance, in the main paper we report DINO-seg using head~4.

We also explore dynamically selecting one box per image among boxes corresponding to the different heads using some heuristics.
We report the two variants that gave the best results. 
In the first variant, we consider selecting the box corresponding to the head with the biggest connected component (`DINO-seg BCC'). However, it yields worse results than with head~4. We also try selecting, over the 6 boxes of the different heads,  the box that has the highest average IoU overlap with the remaining 5 boxes (`DINO-seg HAIoU'). It improves over DINO-seg\,[head~4] by 1 point on both VOC07 and VOC12. However, as shown in \autoref{tab:dinoseg}, it still performs significantly worse than LOST in this single-object discovery task.

\begin{table}[t]
    \small
    \centering
    \begin{tabular}{lcccc}
        \toprule
        Method & VOC07\_trainval & VOC12\_trainval & COCO20k \\
    	\midrule
    	DINO-seg [head 0] & 25.9 & 24.6 & 30.1 \\
    	DINO-seg [head 1] & 36.2 & 35.9 & 35.8 \\
    	DINO-seg [head 2] & 32.1 & 33.2 & 31.6 \\
    	DINO-seg [head 3] & 21.6 & 20.0 & 26.3 \\
    	DINO-seg [head 4] & 45.8 & 46.2 & \bf 42.1 \\
    	DINO-seg [head 5] & 35.5 & 42.1 & 26.5 \\
     	DINO-seg BCC      & 38.8 & 45.2 & 28.8 \\
     	DINO-seg HAIoU    & \bf 46.1 & \bf 47.6 & 40.8 \\
     	\midrule
     	\ours (ours)    & \bf 61.9 & \bf 64.0 & \bf 50.7 \\
    	\bottomrule
    \end{tabular}
    \vspace{2mm}
    \caption{\small \textbf{DINO-seg ablation study.} We compare here CorLoc results on datasets VOC07\_trainval, VOC12\_trainval and COCO20k when applying the DINO-seg method to create a box from the different heads of the attention layer. Also, DINO-seg BCC selects the box/head that produces the biggest connected component, and DINO-seg HAIoU selects the box/head that has the highest average IoU with the other 5 boxes. We additionally report results with our method \ours for comparison.
    }
    \label{tab:dinoseg}
\end{table}

\subsection{Impact of the number of clusters on class-aware detection training}
For the unsupervised class-aware detection experiments of the main paper, we assume that we know the exact number of object classes present in the used dataset, i.e., $20$ in the VOC dataset, and use the same number of K-means clusters.
Here we only assume that we have a rough estimate of the number of classes and study the impact of the requested number of clusters on the performances of the unsupervised detector.

To that end, in \autoref{tab:training-cluster-size}, we provide the mean AP across all the $20$ VOC classes when using $20$, $25$, $30$ and $40$ clusters.
For the case when we use more clusters than the $20$ classes of the VOC dataset,
Hungarian matching, which is used for reporting the AP results, 
will map to the VOC classes only the $20$ most fitted available clusters. 
Thus, when reporting the per-class AP results, we ignore the detections in these unmatched clusters (since they have not been mapped to any ground-truth class).

In \autoref{tab:training-cluster-size}, we observe that our unsupervised detector achieves good results for all the numbers of clusters. Interestingly, for $30$ and $40$ clusters there is a noticeable performance improvement. Similar findings have been observed on prior clustering work~\cite{ji2019invariant, Tian_2021_CVPR, afouras2021self}.

\begin{table}[t]
    \small
    \centering
    \begin{tabular}{c | cccc}
    	\toprule
    	Number $K$ of clusters & $20$ & $25$ & $30$ & $40$\\
     	\midrule
     	Mean AP (\%) & 29.9 & 29.4 & 34.0 & 32.2\\
    	\bottomrule
    \end{tabular}
    \vspace{2mm}
    \caption{\small \textbf{Impact of number of clusters in object detection.} 
    Results, using the mean AP@0.5 (\%) across all the classes, on VOC07 test.
    All models are trained using \ours's pseudo-boxes (i.e., \ours + \od) on the VOC07 and VOC12 trainval sets. The number of classes in VOC is $20$.
    }
    \label{tab:training-cluster-size}
\end{table}

\subsection{Impact of the non-determinism of the K-means clustering}
We investigate the impact of the randomness in the K-means clustering on the results of the object detector. To that end, we repeat 4 times, using different random seeds, the unsupervised class-aware object detection experiment with \ours + OD$^\dagger$ (using the model trained on the union of VOC07 and VOC12 trainval sets, cf.\ \autoref{tab:training-class-ap} in \autoref{sec:results_detection} of the main paper). We obtain a standard deviation of 0.8 for the AP@0.5~\%, which shows that the method is fairly insensitive to the randomness of the clustering method.


\section{More quantitative results and comparisons}

\subsection{Results on more datasets used in previous work} 
For completeness, we present in \autoref{tab:noh} results on the datasets used in previous object discovery works \cite{Vo_2019_CVPR, Wei2019ddtplus, Vo20rOSD, vo2021largescale}. In particular, we evaluate our method on the datasets VOC07$\_$noh and VOC12$\_$noh datasets (also named VOC\_all in literature). They are subsets of the trainval set of the well-known PASCAL VOC 2007 and PASCAL VOC 2012 datasets containing 3550 and 7838 images respectively. These subsets exclude all images containing only objects annotated as ``hard'' or ``truncated'' and all boxes annotated as ``hard'' or ``truncated''.

\begin{table}[t]
    \small
    \centering
    \begin{tabular}{lcc}
    	\toprule
    	Method & VOC07$\_$noh & VOC12$\_$noh\\
    	\midrule
    	OSD \cite{Vo_2019_CVPR}  & 40.7 & - \\
    	DDT+ \cite{Wei2019ddtplus}  &  43.4 & 46.3 \\
    	rOSD \cite{Vo20rOSD} &  49.3 & 51.2 \\
    	LOD \cite{vo2021largescale} & 48.0 & 50.5 \\
    	\ours  & 54.9 & 57.5 \\
    	\bottomrule
    \end{tabular}
    \vspace{5pt}
    \caption{CorLoc results on the VOC07$\_$noh and VOC12$\_$noh datasets.}
    \label{tab:noh}
\end{table}

\subsection{Multi-object discovery results} 
We compare in \autoref{tab:multi_od} the object discovery performance of different methods in the setting where multiple regions are returned per image. This setting has been explored in~\cite{Vo20rOSD} and~\cite{vo2021largescale}.

Following \cite{vo2021largescale}, instead of considering the object recall (detection rate) for a given number of predicted regions per image, as in \cite{Vo20rOSD}, we consider as a metric a form of Average Precision adapted to the task, that we name here ``odAP''. It is the average of the AP of predicted objects for each number of predicted regions, from one to the maximum number of ground-truth objects in an image in the dataset. This odAP metric thus does not depend on the number of detections per image and remains related to AP, which is a standard metric for object detection.
\cite{vo2021largescale} actually uses two variants of this metric: odAP50, where a prediction is correct if its intersection-over-union (IoU) with one of the ground-truth boxes is at least $0.5$, and odAP@[50-95], the average odAP value at 10 equally-spaced values of the IoU threshold between $0.5$ and $0.95$. 

As \ours only returns one region per image, we only consider here \ours + \cad, which is the output of a class-agnostic detector (CAD) trained with \ours boxes, and we compare it to other existing approaches. It can be seen that \ours + \cad outperforms significantly all the previous methods, including the class-agnostic detector trained with LOD~\cite{vo2021largescale} boxes (LOD~\cite{vo2021largescale} + \cad).

\begin{table}[tb]
\vspace{3pt}
    \Large
    \centering
    \resizebox{\textwidth}{!}{%
    \begin{tabular}{lcccccc}
        \toprule
        \multirow{2}{*}{Method} & \multicolumn{3}{c}{odAP50} & \multicolumn{3}{c}{odAP@[50-95]} \\
        \cmidrule{2-7}
         & VOC07\_trainval & VOC12\_trainval & COCO20k\_trainval & VOC07\_trainval & VOC12\_trainval & COCO20k\_trainval \\
        \midrule
        Kim \etal~\cite{kim2009pagerank_uod} & 9.5 & 11.8 & 3.93 & 2.49 & 3.11 & 0.96\\
        DDT+~\cite{Wei2019ddtplus} & 8.7 & 11.1 & 2.41 & 3.0 & 4.1 & 0.73\\
        rOSD~\cite{Vo20rOSD} & 13.1 & 15.4 & 5.18 & 4.29 & 5.27 & 1.62 \\ 
        LOD~\cite{vo2021largescale} & 13.9 & 16.1 & 6.63 & 4.47 & 5.34 & 1.98 \\ 
        LOD~\cite{vo2021largescale} + \cad & 15.8 & 20.9 & 7.26 & 5.03 & 7.07 & 2.28 \\
        \ours + \cad & \textbf{19.8} & \textbf{24.9} & \textbf{7.93} & \textbf{6.71} & \textbf{8.85} & \textbf{2.51} \\
        \bottomrule
    \end{tabular}
    }
    \vspace{-5pt}
    \caption{\small Multi-object discovery performance in odAP (Average Precision for object discovery) of our method and the baselines \cite{kim2009pagerank_uod, Wei2019ddtplus, Vo20rOSD, vo2021largescale}.}
    \label{tab:multi_od}
\end{table}

\subsection{Image nearest neighbor retrieval}
Following LOD~\cite{vo2021largescale}, we use \ours box descriptors to find images that are similar to each other (image neighbors) in the image collection.

To this end, each image is represented by the \cls descriptors of its \ours box and the cosine similarity between these descriptors is used to define a similarity between the images. Then, for each image, the top $\tau$ images with the highest similarity are chosen as its neighbors. Similar to LOD~\cite{vo2021largescale}, we choose $\tau=10$ and use CorRet~\cite{Cho_2015_CVPR} as the evaluation metric, defined as the average percentage of the
retrieved image neighbors that are actual neighbors (i.e., that contain objects of the same category) in the ground-truth image graph over all images. 

We compare the performance of our method in this task with rOSD~\cite{Vo20rOSD} and LOD~\cite{vo2021largescale} in \autoref{tab:corret}. We see that \ours boxes, when represented by DINO~\cite{caron2021emerging} features, yield the better CorRet score compared to~\cite{Vo20rOSD,vo2021largescale}. When VGG16~\cite{Symonian2014verydeep} features are used, \ours is behind LOD~\cite{vo2021largescale} but better than rOSD~\cite{Vo20rOSD}. 

\begin{table}[htb]
\vspace{3pt}
\parbox{0.32\linewidth}{
\small
\centering
\vspace{3pt}
\resizebox{0.32\textwidth}{!}{%
\begin{tabular}{lcc}
    \toprule
    Method & Features & CorRet (\%) \\
    \midrule
    rOSD~\cite{Vo20rOSD} & VGG16~\cite{Symonian2014verydeep} & 64 \\
    LOD~\cite{vo2021largescale} & VGG16~\cite{Symonian2014verydeep} & 70 \\ 
    \ours (ours) & VGG16~\cite{Symonian2014verydeep} & 68 \\
    \ours (ours) & DINO~\cite{caron2021emerging} & \bf 72 \\
    \toprule
\end{tabular}
}
\vspace{-7pt}
\caption{\small Image neighbor retrieval performance (CorRet) of different methods.}
\label{tab:corret}
}
\hfill
\parbox{.65\linewidth}{
\small
\centering
\resizebox{0.65\textwidth}{!}{%
\begin{tabular}{lcccc}
    \toprule
    \multirow{2}{*}{Method} & \multirow{2}{*}{Features} & \multicolumn{3}{c}{CorLoc (\%)} \\
    \cmidrule{3-5}
     & & VOC07\_trainval & VOC12\_trainval & COCO20k\_trainval \\
    \midrule
    LOD~\cite{vo2021largescale} & VGG16~\cite{Symonian2014verydeep} & 53.6 & 55.1 & 48.5 \\
    LOD~\cite{vo2021largescale} & DINO~\cite{caron2021emerging} & 43.2 & 45.9 & 33.7 \\
    \ours (ours) & VGG16~\cite{Symonian2014verydeep} & 42.0 & 47.2 & 30.2 \\
    \ours (ours) & DINO~\cite{caron2021emerging} & \bf 61.9 & \bf 64.0 & \bf 50.7 \\
    \bottomrule
\end{tabular}
}
\vspace{-5pt}
\caption{\small Single-object discovery performance in CorLoc of LOD~\cite{vo2021largescale} and \ours with different types of features.}
\label{tab:corloc_different_features}
}
\end{table}

\subsection{Using DINO features} 
We are aware that, in \autoref{tab:training} of the main paper, we compare our method using a transformer backbone to methods based on a VGG16 pre-trained on ImageNet models. For a fair comparison, we investigate here the state-of-the-art LOD~\cite{vo2021largescale} method when adapted to use the transformers features. 

LOD~\cite{vo2021largescale} uses the algorithm from rOSD~\cite{Vo20rOSD} to generate region proposals from CNN features for their pipeline, but we observe that this algorithm does not yield good proposals with transformer features. We therefore run LOD with edgeboxes~\cite{zitnick2014edge} and use DINO~\cite{caron2021emerging} features, extracted with ROIPool~\cite{girshick2015ICCV_fast_rcnn}, to represent these proposals. We present the results on VOC07\_trainval, VOC12\_trainval and COCO20k\_trainval dataset in \autoref{tab:corloc_different_features}.

Our results in \autoref{tab:backbone} of the main paper show that a direct adaption of \ours, designed by analysing the properties of transformers features, to CNN features yields worse performance. Conversely, as we see in~\autoref{tab:corloc_different_features} here, adapting algorithms developed using properties of CNN features to transformer features is also not direct. Nevertheless, the number of design choices to adapt these algorithms to new types of features is vast and we do not exclude that some design choices might improve the results even further, e.g., by exploiting together CNN and transformer features.

\subsection{Using supervised pre-training.}
We test \ours but this time using a transformer pre-trained under full supervision on ImageNet. We use the model provided by DeiT~\cite{touvron2020deit}. 

With this model, \ours achieves a CorLoc of $16.9$\% which is significantly worse than the results obtained with the DINO self-supervised pre-trained model. We remark that a similar observation was made for DINO~\cite{caron2021emerging}, where the segmentation performance obtained with the model trained under full supervision yields significantly worse results than when using DINO's model. It is unclear, however, if this difference of performance can be attributed to the properties of the self-supervision loss or to the more aggressive data augmentation used during DINO pre-training.

\section{More visualizations (single- and multi-object discovery)}

\begin{figure*}[t]
\centering
\begin{minipage}{.39\linewidth}
\centering
    \includegraphics[height=34mm]{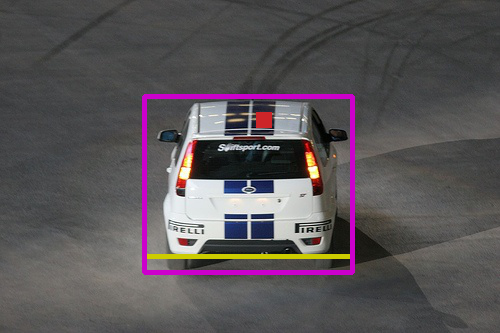}\\
    \includegraphics[height=34mm,trim={1mm 0 0 0},clip]{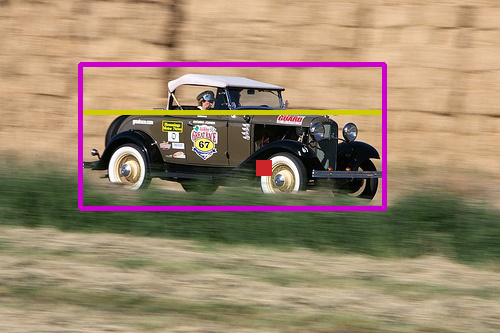}\\
    \includegraphics[height=34mm,trim={10mm 0 0 0},clip]{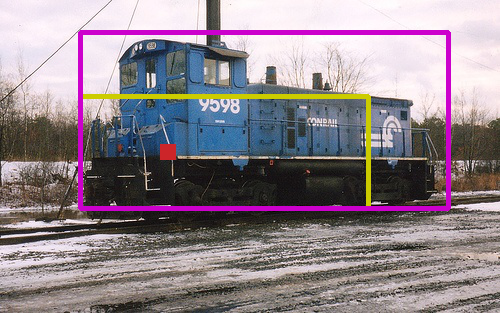}\\
    \includegraphics[height=34mm,trim={25mm 0 0 0},clip]{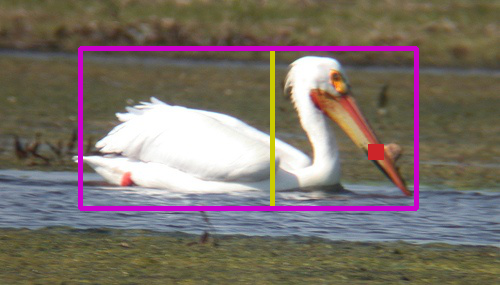}\\
    \includegraphics[height=34mm,trim={0 0 0 0},clip]{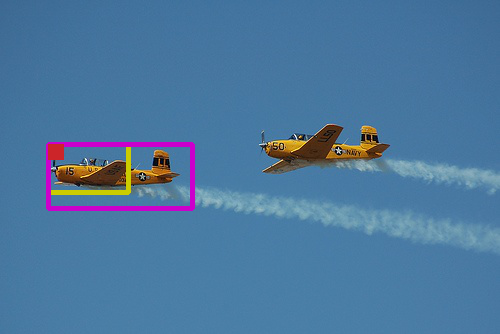}
\end{minipage}
\begin{minipage}{.39\linewidth}
\centering
    \includegraphics[height=34mm,trim={0mm 0 0mm 0},clip]{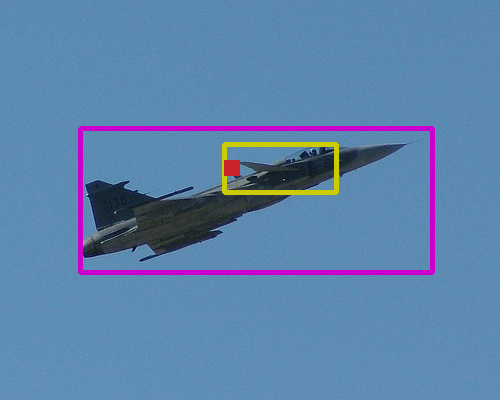}\\
    \includegraphics[height=34mm]{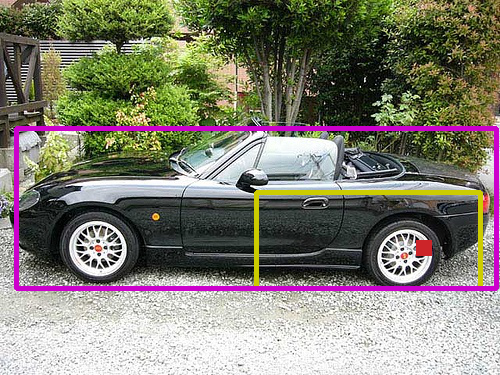}\\
    \includegraphics[height=34mm,trim={25mm 0 0 0},clip]{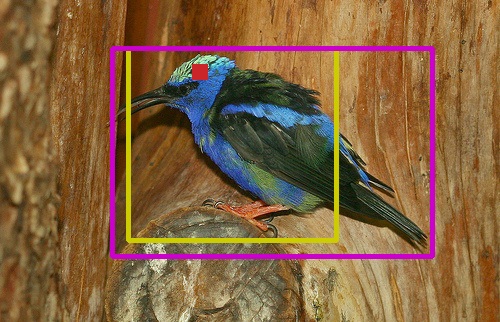}\\
    \includegraphics[height=34mm,trim={20mm 0 0 0},clip]{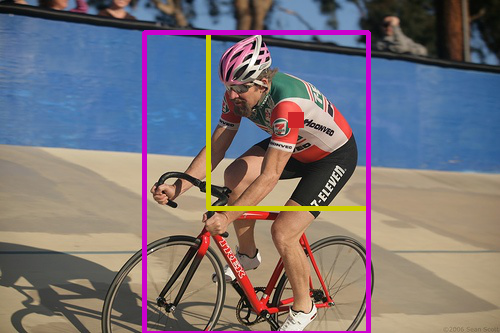}\\
    \includegraphics[height=34mm,trim={14mm 0 6mm 0},clip]{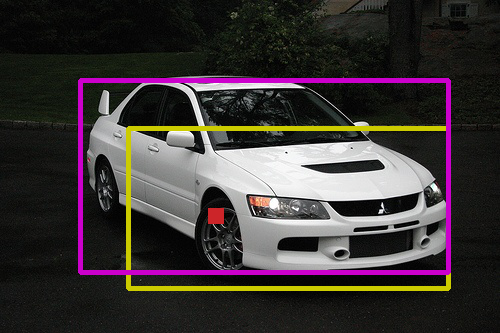}
\end{minipage}
\begin{minipage}{.19\linewidth}
\centering
    \includegraphics[height=34mm]{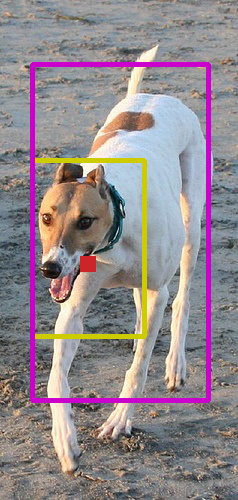}\\
    \includegraphics[height=34mm,trim={12mm 0 4mm 0},clip]{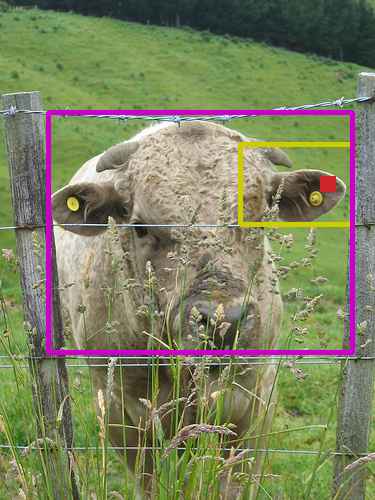}\\
    \includegraphics[height=34mm]{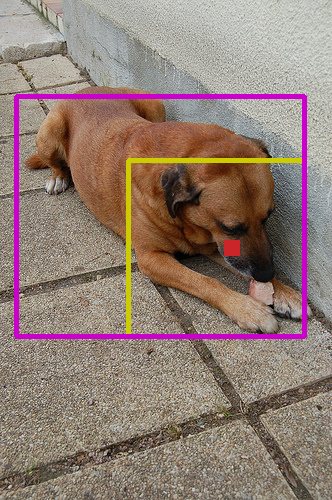}\\
    \includegraphics[height=34mm]{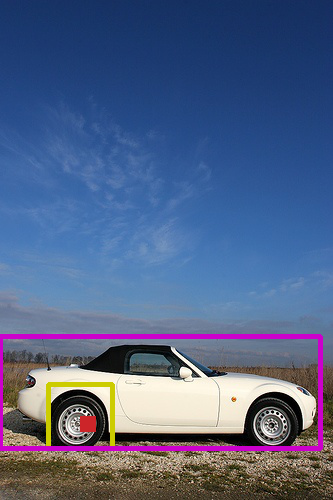}\\
    \includegraphics[height=34mm]{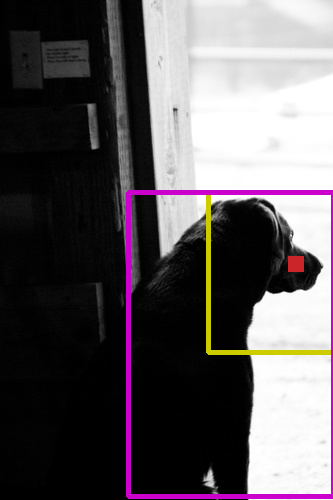}
\end{minipage}
\vspace{5pt}
\caption{\textbf{Object localization on VOC07.} The red square represents the seed $p^*$, the yellow bos is the box obtained using only the seed $p^*$, and the purple box is the box obtained using all the seeds $\mathcal{S}$ with $k=100$.}
\label{fig:seed_expansion}
\end{figure*}

\begin{figure*}[t]
\centering
\begin{minipage}{.32\linewidth}
\centering
    \includegraphics[height=28mm,trim={0mm 0 0 0},clip]{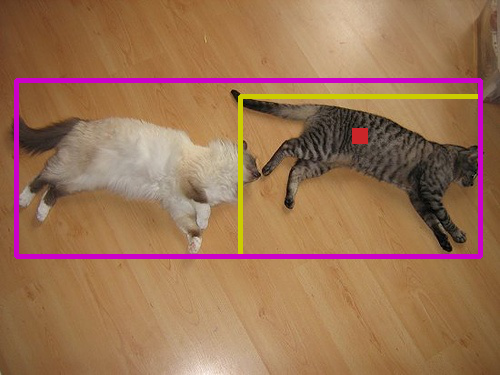}\\
    \includegraphics[height=28mm,trim={0mm 0 0 0},clip]{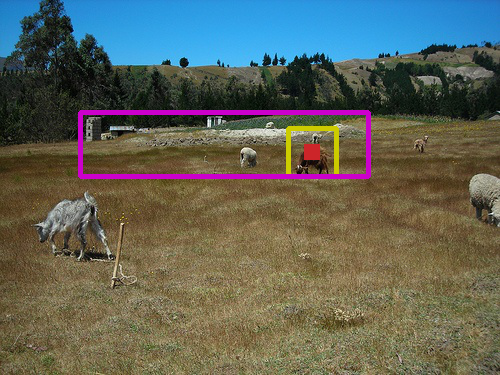}\\
\end{minipage}
\begin{minipage}{.32\linewidth}
\centering
    \includegraphics[height=28mm,trim={0 0 0 0},clip]{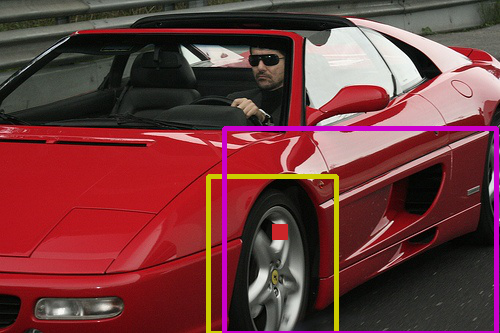}
    \includegraphics[height=28mm,trim={0 0 0 0},clip]{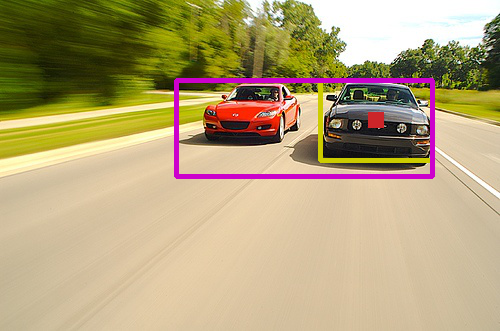}
\end{minipage}
\begin{minipage}{.32\linewidth}
\centering
    \includegraphics[height=28mm,trim={20mm 0 0 0},clip]{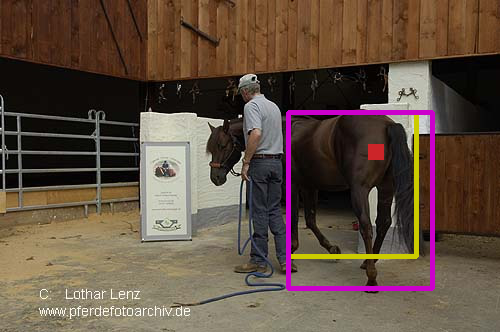}
    \includegraphics[height=28mm,trim={0 0 0 0},clip]{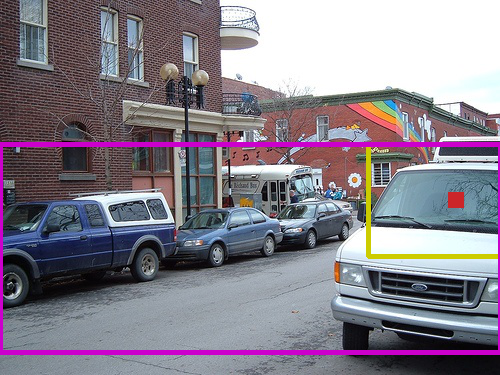}
\end{minipage}
\vspace{5pt}
\caption{\textbf{Cases of localization failure on VOC07.} The red square represents the seed $p^*$, the yellow box is the box obtained using only the seed $p^*$, and the purple box is the box obtained using all the seeds $\mathcal{S}$ with $k=100$.}
\label{fig:seed_expansion_failure}
\end{figure*}

\begin{figure*}[t]
\centering
\begin{minipage}{.32\linewidth}
\centering
    \includegraphics[height=28mm,trim={0mm 0 0 0},clip]{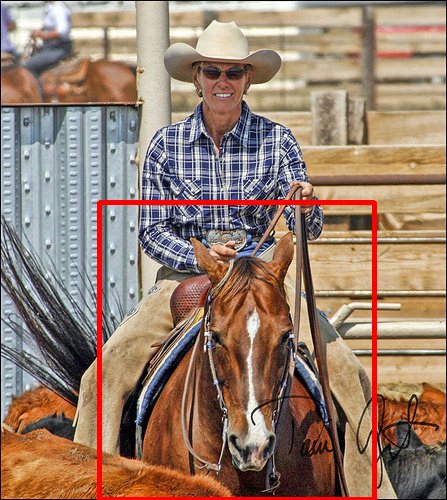}\\
    \includegraphics[height=28mm,trim={0mm 0 0 0},clip]{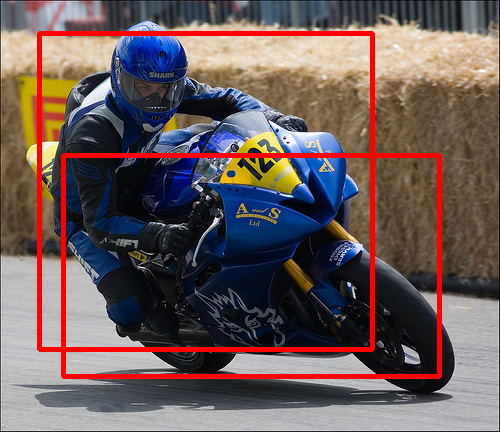}\\
\end{minipage}
\begin{minipage}{.32\linewidth}
\centering
    \includegraphics[height=28mm,trim={0 0 0 0},clip]{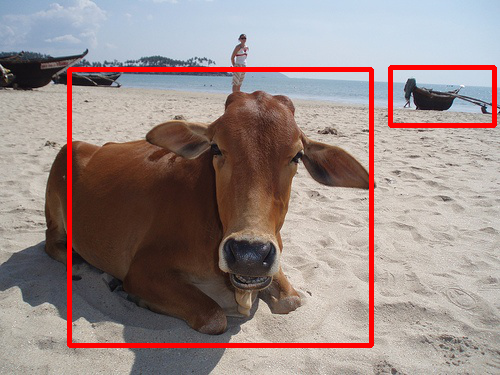}\\
    \includegraphics[height=28mm,trim={0 0 0 0},clip]{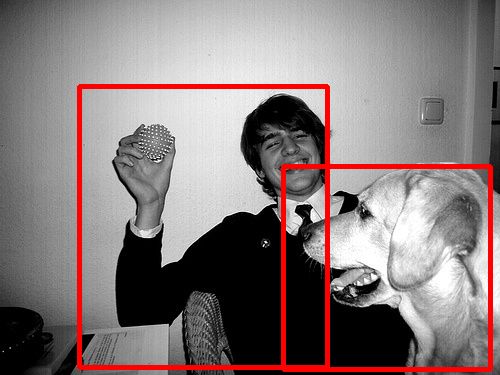}
\end{minipage}
\begin{minipage}{.32\linewidth}
\centering
    \includegraphics[height=28mm,trim={0 0 0 0},clip]{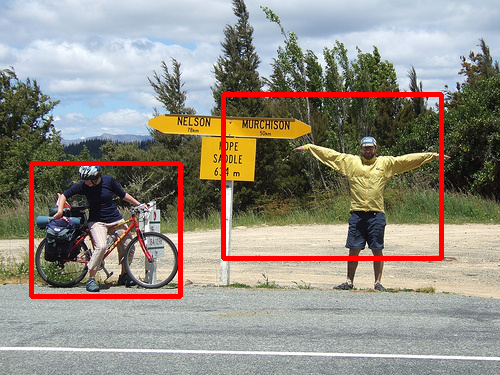}\\
    \includegraphics[height=28mm,trim={0 0 0 0},clip]{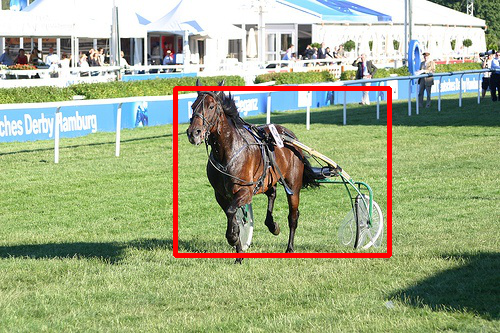}
\end{minipage}
\begin{minipage}{.32\linewidth}
\centering
    \includegraphics[height=28mm,trim={0 0 0 0},clip]{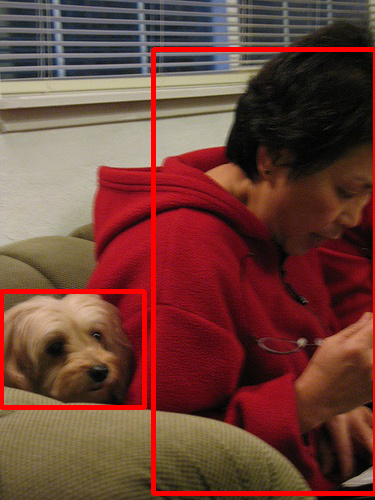}\\
    \includegraphics[height=28mm,trim={0 0 0 0},clip]{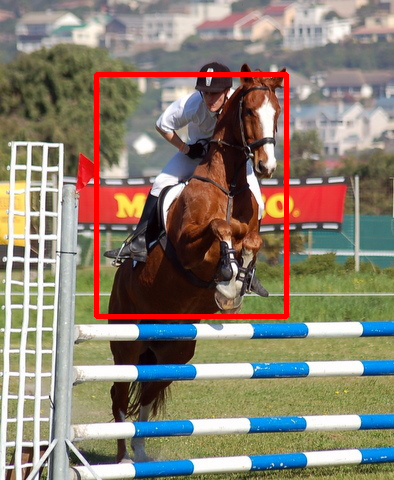}
\end{minipage}
\begin{minipage}{.32\linewidth}
\centering
    \includegraphics[height=28mm,trim={0 0 0 0},clip]{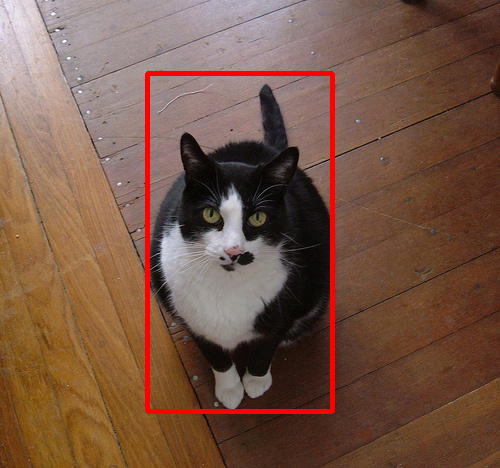}\\
    \includegraphics[height=28mm,trim={0 0 0 0},clip]{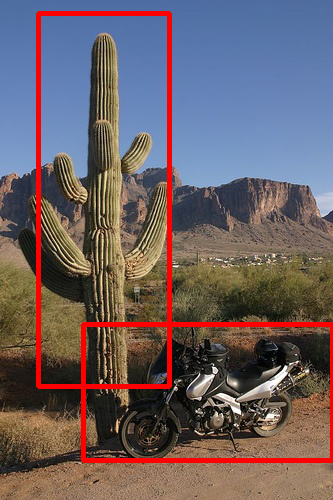}
\end{minipage}
\begin{minipage}{.32\linewidth}
\centering
    \includegraphics[height=28mm,trim={0 0 0 0},clip]{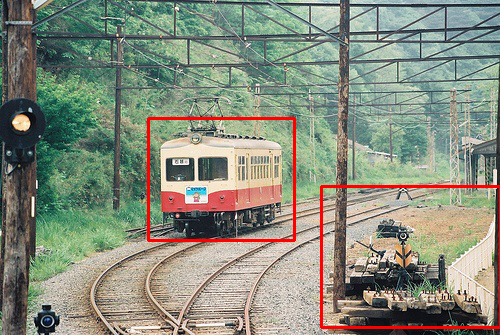}\\
    \includegraphics[height=28mm,trim={0 0 0 0},clip]{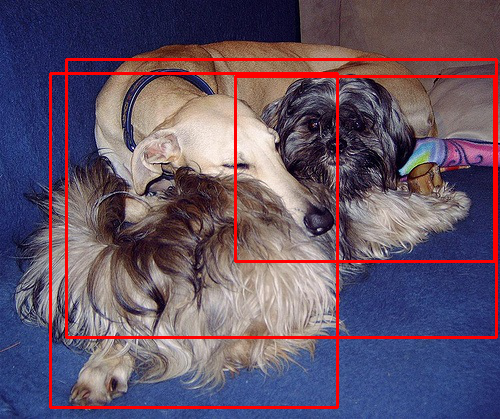}
\end{minipage}
\vspace{5pt}
\caption{\textbf{Multi-object discovery on VOC07 (\ours + \cad).} Predictions performed by the class-agnostic detector on VOC07.}
\label{fig:multi-voc}
\end{figure*}

\begin{figure*}[t]
\centering
\begin{minipage}{.23\linewidth}
\centering
    \includegraphics[height=24mm,trim={50mm 0 30mm 0},clip]{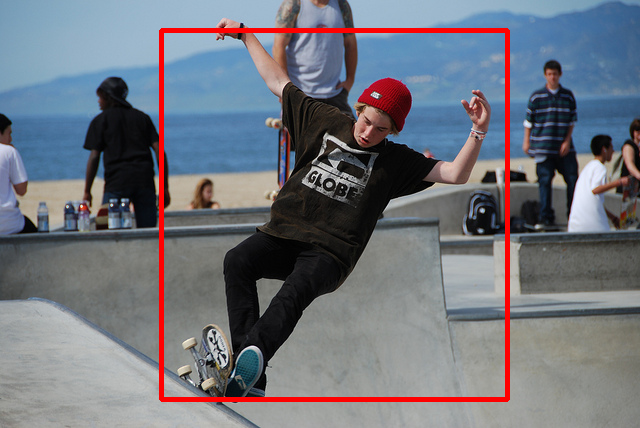}\\
    \includegraphics[height=24mm,trim={50mm 0 30mm 0},clip]{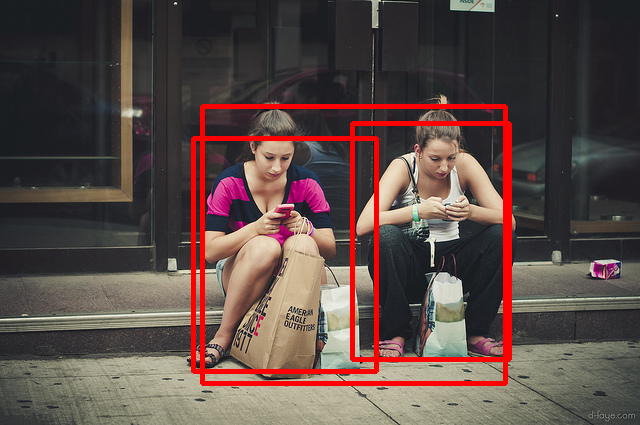}
\end{minipage}
\begin{minipage}{.25\linewidth}
\centering
    \includegraphics[height=24mm,trim={0 0 15mm 0},clip]{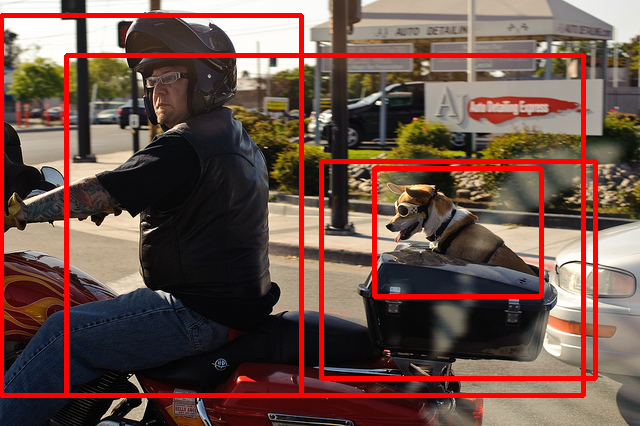}\\
    \includegraphics[height=24mm,trim={0 0 0 0},clip]{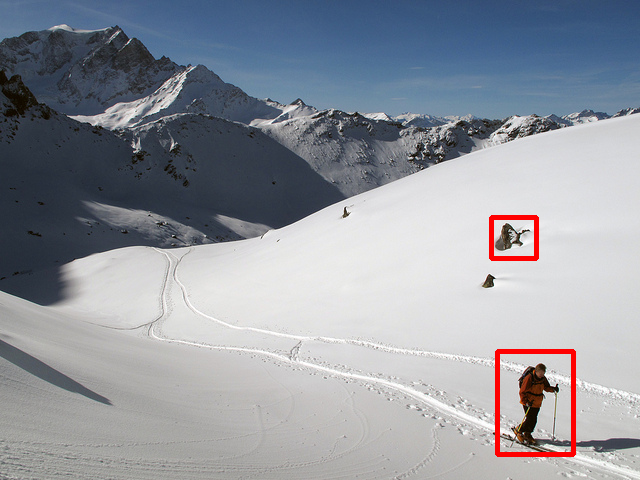}
\end{minipage}
\begin{minipage}{.23\linewidth}
\centering
    \includegraphics[height=24mm,trim={20mm 0 0 0},clip]{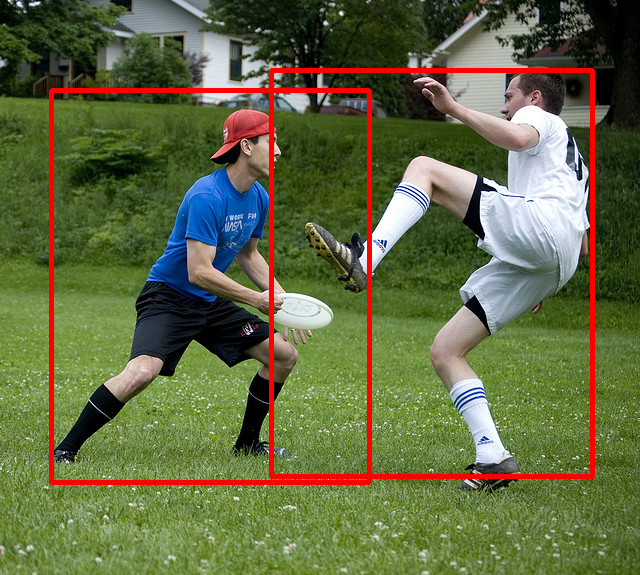}\\
    \includegraphics[height=24mm,trim={0 0 0 0},clip]{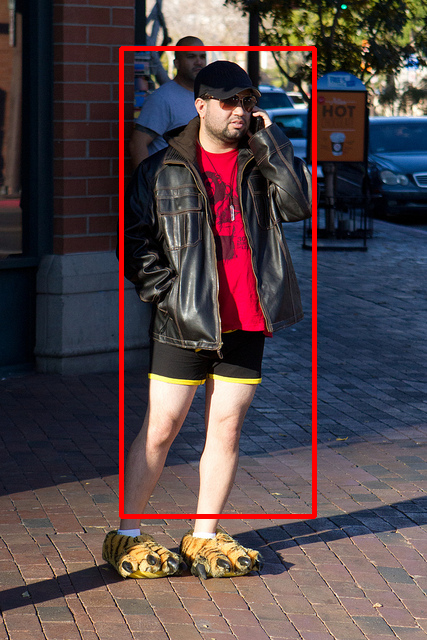}
\end{minipage}
\begin{minipage}{.21\linewidth}
\centering
    \includegraphics[height=24mm,trim={0 0 0 0},clip]{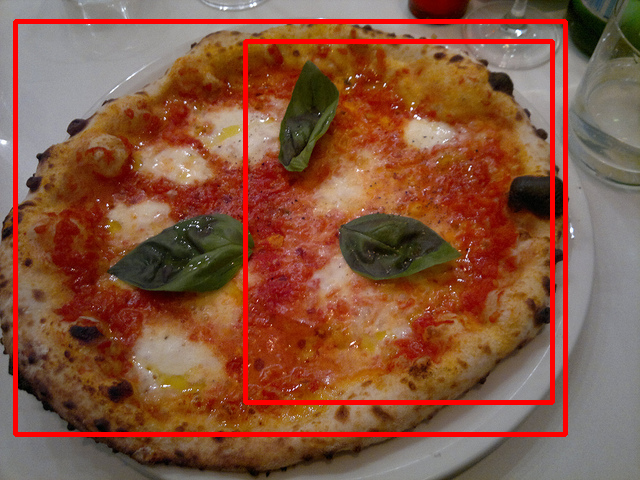}\\
    \includegraphics[height=24mm,trim={0 0 0 0},clip]{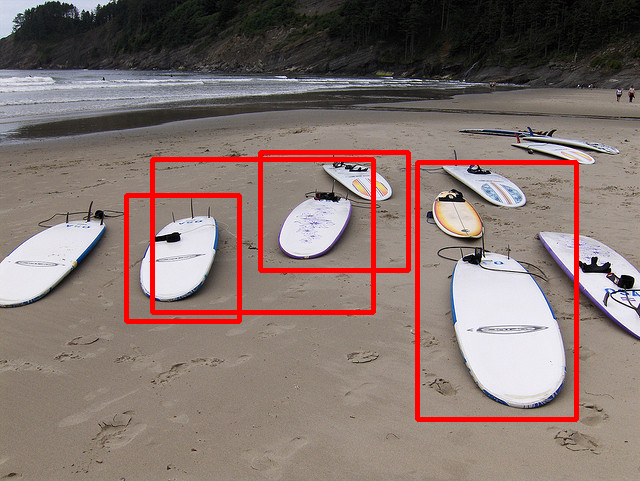}
\end{minipage}
\vspace{5pt}
\caption{\textbf{Multi-object discovery on COCO (\ours + \cad).} Predictions performed by the class-agnostic detector on COCO.}
\label{fig:multi-coco}
\end{figure*}

\begin{figure*}[t]
\centering
\begin{minipage}{.32\linewidth}
\centering
    \includegraphics[height=28mm,trim={0mm 0 0 0},clip]{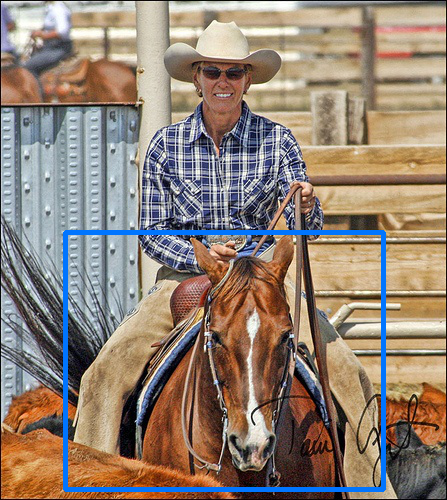}\\
    \includegraphics[height=28mm,trim={0mm 0 0 0},clip]{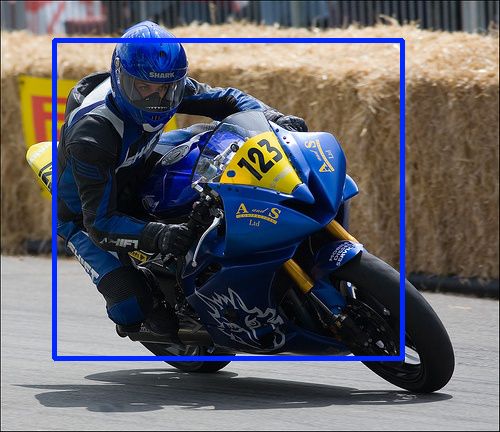}\\
\end{minipage}
\begin{minipage}{.32\linewidth}
\centering
    \includegraphics[height=28mm,trim={0 0 0 0},clip]{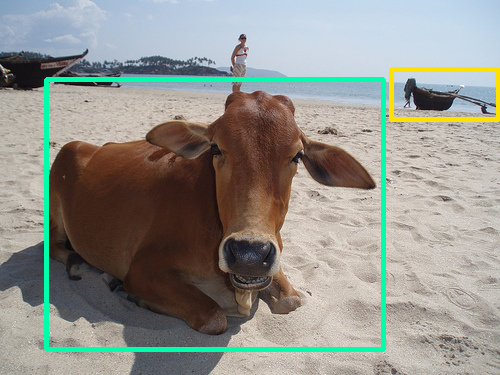}\\
    \includegraphics[height=28mm,trim={0 0 0 0},clip]{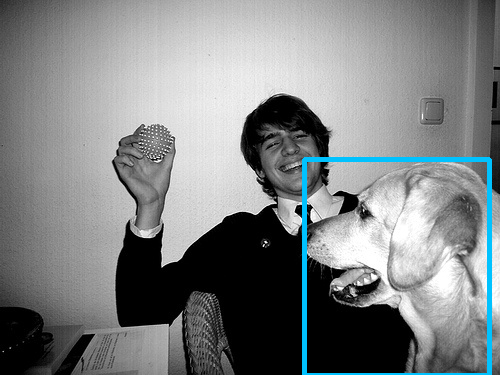}
\end{minipage}
\begin{minipage}{.32\linewidth}
\centering
    \includegraphics[height=28mm,trim={0 0 0 0},clip]{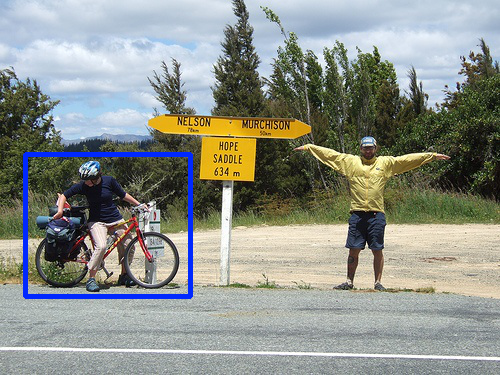}\\
    \includegraphics[height=28mm,trim={0 0 0 0},clip]{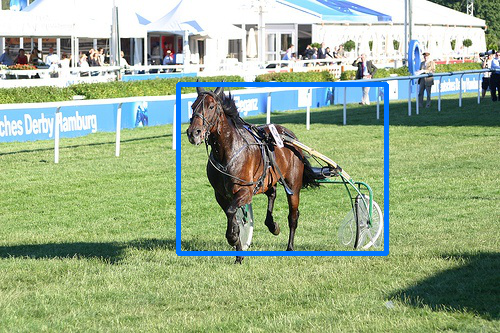}
\end{minipage}
\begin{minipage}{.32\linewidth}
\centering
    \includegraphics[height=28mm,trim={0 0 0 0},clip]{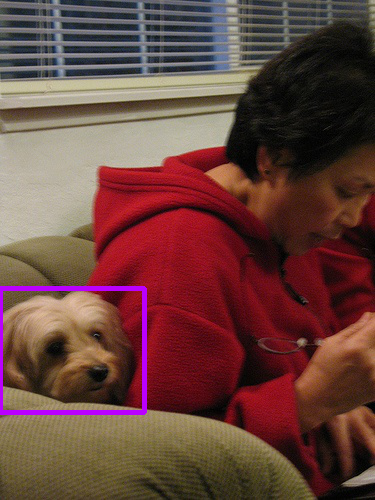}\\
    \includegraphics[height=28mm,trim={0 0 0 0},clip]{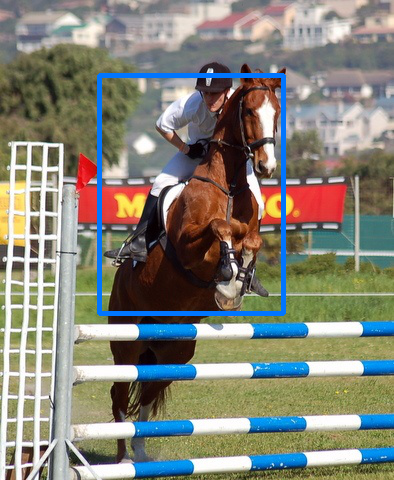}
\end{minipage}
\begin{minipage}{.32\linewidth}
\centering
    \includegraphics[height=28mm,trim={0 0 0 0},clip]{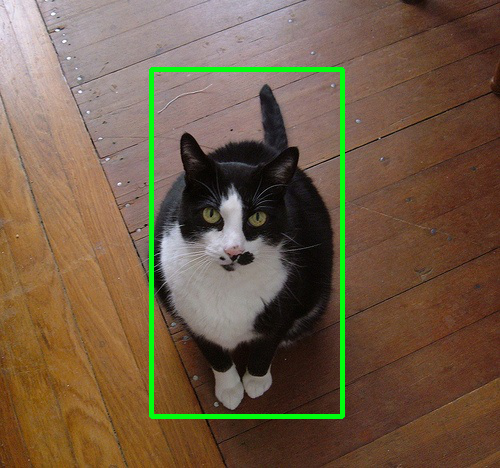}\\
    \includegraphics[height=28mm,trim={0 0 0 0},clip]{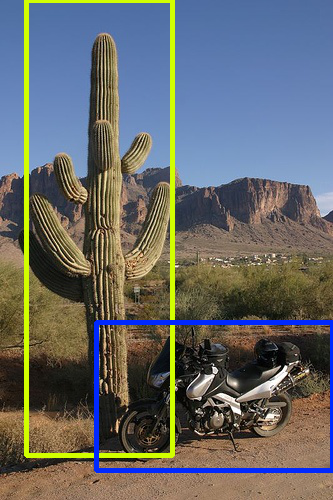}
\end{minipage}
\begin{minipage}{.32\linewidth}
\centering
    \includegraphics[height=28mm,trim={0 0 0 0},clip]{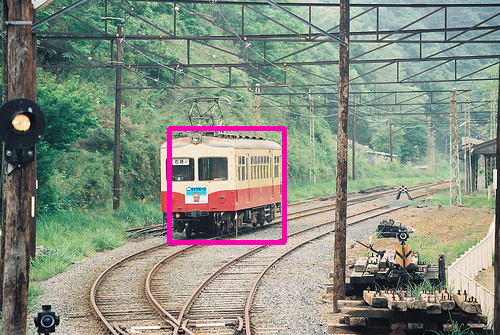}\\
    \includegraphics[height=28mm,trim={0 0 0 0},clip]{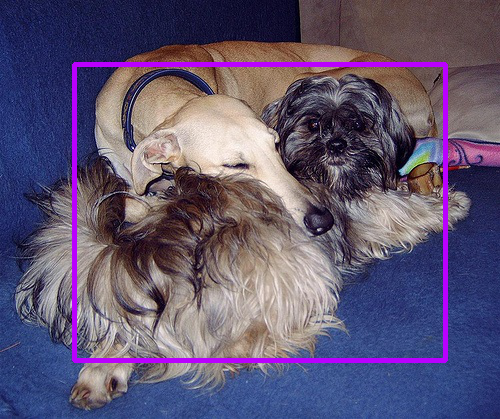}
\end{minipage}
\vspace{5pt}
\caption{\textbf{Multi-object discovery on VOC07 (\ours + \od).} Predictions performed by the class-aware detector on VOC07 (a different color per class).}
\label{fig:multi-od-voc}
\end{figure*}

\begin{figure*}[t]
\centering
\begin{minipage}{.23\linewidth}
\centering
    \includegraphics[height=24mm,trim={50mm 0 30mm 0},clip]{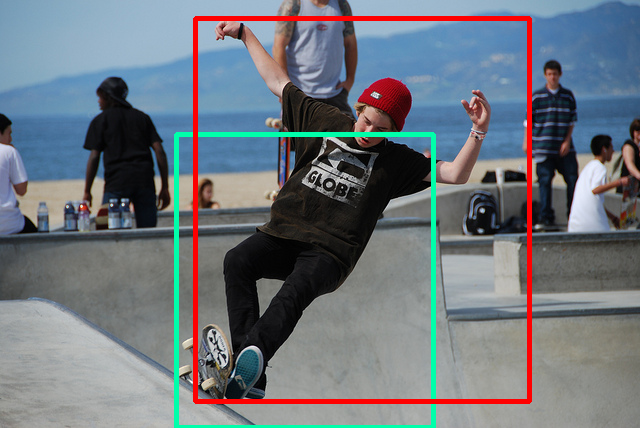}\\
    \includegraphics[height=24mm,trim={50mm 0 30mm 0},clip]{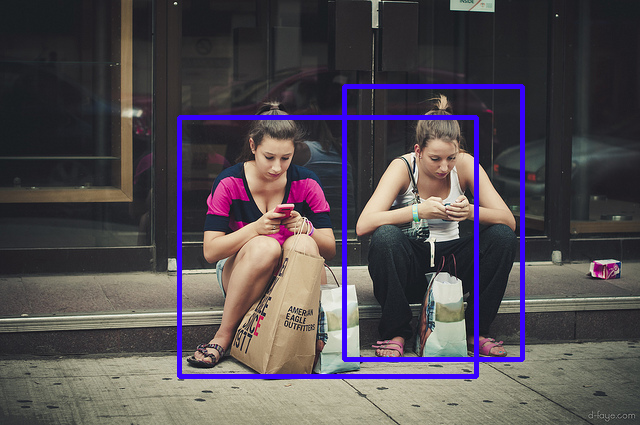}
\end{minipage}
\begin{minipage}{.25\linewidth}
\centering
    \includegraphics[height=24mm,trim={0 0 15mm 0},clip]{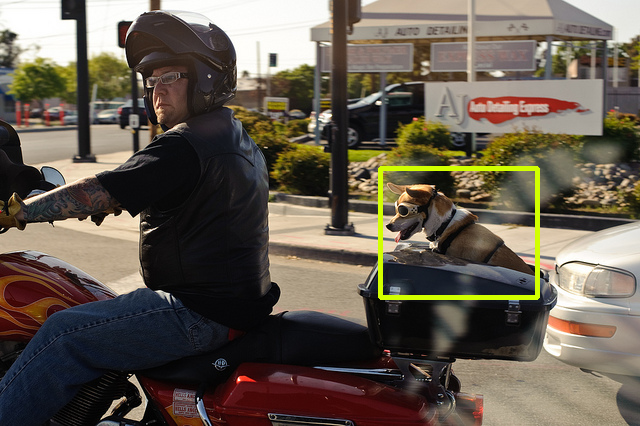}
    \includegraphics[height=24mm,trim={0 0 0 0},clip]{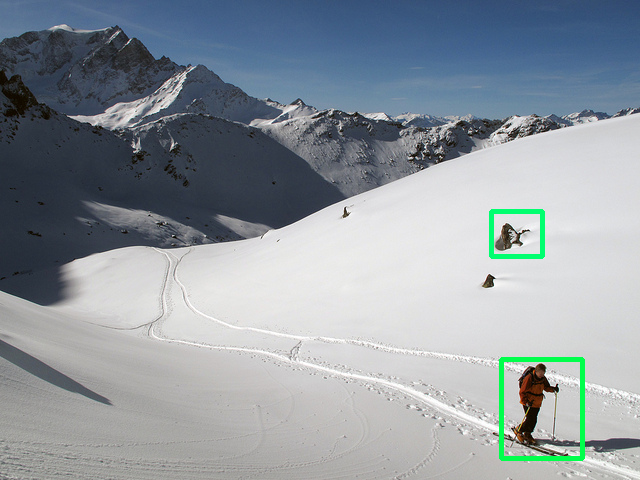}
\end{minipage}
\begin{minipage}{.23\linewidth}
\centering
    \includegraphics[height=24mm,trim={20mm 0 0 0},clip]{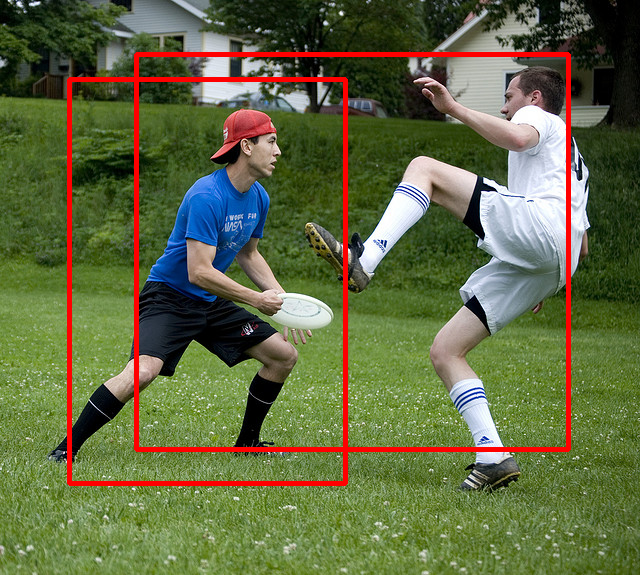}
    \includegraphics[height=24mm,trim={0 0 0 0},clip]{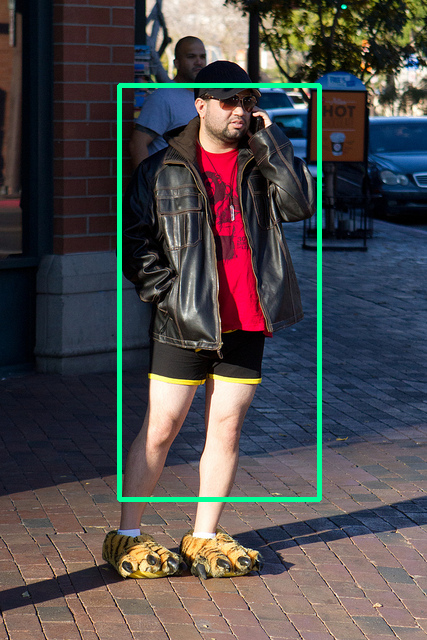}
\end{minipage}
\begin{minipage}{.21\linewidth}
\centering
    \includegraphics[height=24mm,trim={0 0 0 0},clip]{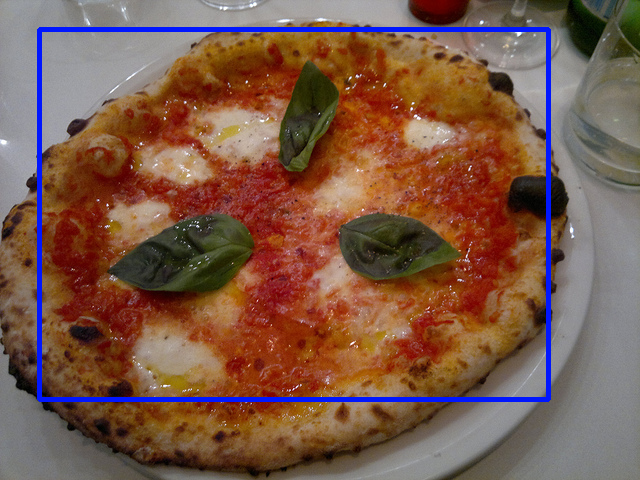}
    \includegraphics[height=24mm,trim={0 0 0 0},clip]{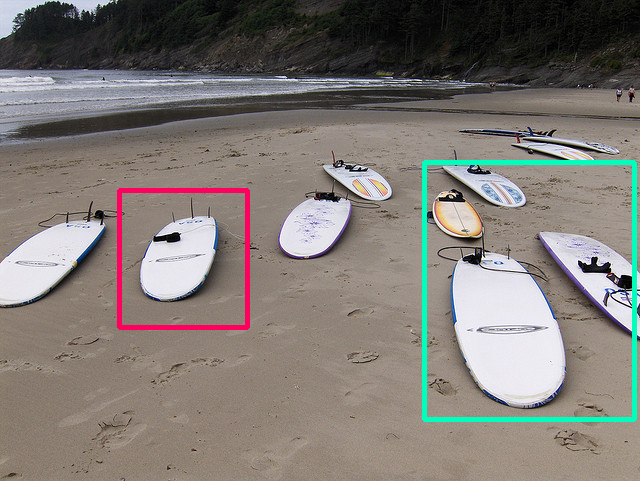}
\end{minipage}
\vspace{5pt}
\caption{\textbf{Multi-object discovery on COCO (\ours + \od).} Predictions performed by the class-aware detector on COCO (a different color per class). The actual ``person'' class is assigned three different pseudo-classes, illustrating the difficulty to see a single category for a ``person'' in very different positions.}
\label{fig:multi-od-coco}
\end{figure*}

We present in Figures~\ref{fig:seed_expansion}-\ref{fig:multi-od-coco} additional qualitative results of our method.

\autoref{fig:seed_expansion} and \autoref{fig:seed_expansion_failure} are discussed in the \autoref{sec:expansion}. 

\autoref{fig:multi-voc} and \autoref{fig:multi-coco} show successful examples of \ours + \cad in VOC07\_trainval and COCO20k\_trainval datasets. It can be seen that it is able to localize multiple objects in the same image. 

\autoref{fig:multi-od-voc} and \autoref{fig:multi-od-coco} present results obtained with {\ours} + \od on the VOC07 and COCO datasets respectively. They show the localization predictions with their predicted pseudo-classes. Each pseudo-class is assigned a different color. In  \autoref{fig:multi-od-coco}, the ``person" objects are assigned three different pseudo-classes; those failures show the difficulty to assign the same class to ``person'' in very different positions. 

\section{Training details of the Faster R-CNN detection models}

In the main paper, we explore the application of \ours in unsupervised object detection by using its pseudo-boxes as ground truth for training Faster R-CNN detection models. 

For the implementation of the Faster R-CNN detector,
we use the \texttt{R50-C4} model of Detectron2~\cite{wu2019detectron2} that relies on a ResNet-50 \cite{He2016cvpr_resnet} backbone.
In our experiments, this ResNet-50 backbone is pre-trained with DINO self-supervision.
Then, to train the Faster R-CNN model on the considered dataset, we use the protocol and most hyper-parameters from He~\etal~\cite{he2020momentum}. 

In details, we train with mini-batches of size $16$ across $8$ GPUs using \texttt{SyncBatchNorm} to finetune BatchNorm parameters, as well as adding an extra BatchNorm layer for the RoI head after \texttt{conv5}, i.e., \texttt{Res5ROIHeadsExtraNorm} layer in Detectron2.
During training, the learning rate is first warmed-up for $100$ steps to $0.02$ and then reduced by a factor of $10$ after $18$K and $22$K training steps.
We use in total $24$K training steps for all the experiments, except when training class-agnostic detectors on the pseudo-boxes of the VOC07 trainval set, in which case we use $10$K steps.
For all experiments, during training, we freeze the first two convolutional blocks of ResNet-50, i.e., \texttt{conv1} and \texttt{conv2} in Detectron2.

\end{document}